%% file: acl_latex.tex
\documentclass[11pt]{article}

\usepackage[final]{acl}

\usepackage{times}
\usepackage{latexsym}

\usepackage[T1]{fontenc}

\usepackage[utf8]{inputenc}

\usepackage{microtype}

\usepackage{inconsolata}

\usepackage{graphicx}
\usepackage{amssymb}

\usepackage{array}
\usepackage{booktabs}
\usepackage{multirow}
\usepackage{rotating}

\usepackage[table]{xcolor}

\usepackage{tcolorbox}
\tcbuselibrary{listings,skins,breakable}

\newtcolorbox{promptbox}[1][]{
  enhanced,
  breakable,
  colback=gray!5,
  colframe=gray!50,
  fonttitle=\bfseries\ttfamily,
  title={#1},
  left=2mm,
  right=2mm,
  top=1mm,
  bottom=1mm,
  fontupper=\small\ttfamily,
  before upper={\setlength{\parindent}{0pt}},
}

\title{SkMTEB: Slovak Massive Text Embedding Benchmark\\ and Model Adaptation}

\author{\bf Marek \v{S}uppa$^{\alpha,~\beta}$\thanks{\,Correspondence: \href{mailto:marek@suppa.sk}{marek@suppa.sk}} \quad Andrej Ridzik$^{\delta}$ \quad Daniel Hl\'adek$^{\gamma}$\\
\bf Nat\'{a}lia K\v{n}a\v{z}ekov\'{a}$^{\alpha,~\delta}$ \quad Vikt\'{o}ria Ondrejov\'{a}$^{\beta}$\\
$^{\alpha}$Comenius University in Bratislava, Slovakia,
$^{\beta}$Cisco Systems, \\
$^{\gamma}$Technical University of Ko\v{s}ice, Slovakia, \\
$^{\delta}$Kempelen Institute of Intelligent Technologies, Bratislava, Slovakia \\[4pt]
{\small \href{https://github.com/slovak-nlp/skmteb}{\raisebox{-0.2em}{\includegraphics[height=1em]{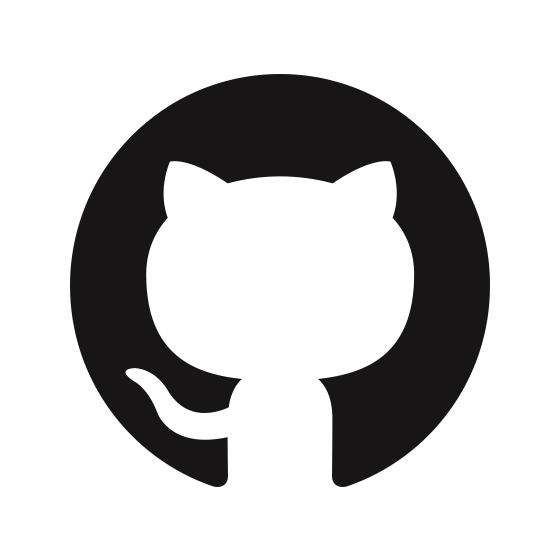}}~\texttt{github.com/slovak-nlp/skmteb}}}\\
{\small \href{https://huggingface.co/collections/slovak-nlp/skmteb}{\raisebox{-0.2em}{\includegraphics[height=1em]{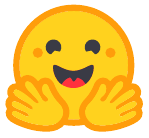}}~\texttt{huggingface.co/collections/slovak-nlp/skmteb}}}\\[10pt]
}

\begin{document}
\maketitle

\begin{abstract}
We introduce SkMTEB, the first comprehensive MTEB-style text embedding benchmark for Slovak, a low-resource West Slavic language, comprising 31 datasets across 7 task types---nearly 4$\times$ the depth of existing multilingual benchmark coverage for Slovak. Our evaluation of 31 embedding models reveals that large instruction-tuned multilingual models achieve the strongest performance, while existing Slovak-specific models trained for NLU tasks transfer poorly to embedding tasks. To address the need for efficient, locally-deployable Slovak embeddings, we develop \texttt{e5-sk-small} (45M parameters) and \texttt{e5-sk-large} (365M) by applying vocabulary trimming and fine-tuning to Multilingual E5 models. Despite size reductions of up to 62\%, our open-source models achieve competitive performance with proprietary APIs while remaining locally deployable for semantic search and retrieval-augmented generation (RAG). We release the benchmark, models, datasets, and code openly, hoping our approach offers a replicable path for other under-resourced languages.
\end{abstract}

\section{Introduction}

Text embeddings are now core infrastructure for semantic search, retrieval-augmented generation (RAG), clustering, and classification. The field has pursued scale---with state-of-the-art models reaching billions of parameters---but benchmark evidence is concentrated in high-resource languages, and the most capable models remain impractical to deploy at low latency or on constrained hardware.

This tension is especially acute for under-resourced languages. Although large multilingual models technically support hundreds of languages, their capacity is predominantly allocated to high-resource languages like English and Chinese. For a language like Slovak -- a West Slavic language with approximately 5 million speakers---this means suboptimal representation in model vocabularies, limited training data coverage, and ultimately degraded performance compared to well-resourced languages.

Two complementary developments are needed to address this gap. First, robust evaluation benchmarks are essential to measure progress and identify weaknesses in the Slovak embedding models. Although benchmarks such as MTEB~\cite{muennighoff-etal-2023-mteb} have catalyzed embedding research for English, and language-specific benchmarks have emerged for Chinese~\cite{xiao-etal-2023-cpack}, Polish~\cite{poswiata-etal-2024-plmteb} and other languages, Slovak lacks such evaluation infrastructure. The existing skLEP benchmark~\cite{suppa-etal-2025-sklep} addresses natural language understanding but not the embedding tasks critical to retrieval and semantic similarity applications.

Second, efficient adaptation techniques are needed to create compact and performant models that can be used practically. Approaches such as vocabulary trimming~\cite{ushio-etal-2023-vocab-trimming} and targeted fine-tuning on high-quality data can yield strong results without massive compute budgets. For under-resourced languages, the goal is not to match the largest models on general benchmarks, but to create practical, efficient models that serve the specific language well.

In this work, we address both needs for Slovak. We introduce \textbf{SkMTEB}, the first comprehensive Slovak text embedding benchmark, comprising 31 datasets across 7 task types. Beyond evaluation, we demonstrate that effective Slovak embedding models can be trained with relatively modest resources by fine-tuning existing models on curated Slovak data and applying vocabulary trimming to create compact, language-specific variants.

Our contributions can hence be summarized as follows:

\begin{itemize}
    \item We introduce SkMTEB---the first Slovak Massive Text Embedding Benchmark, for which we collate 31 datasets across 7 diverse task types.
    \item To ensure the benchmark has sufficient breadth, despite the severe lack of datasets in Slovak, we adapt multiple existing datasets for new tasks while also introducing seven brand-new datasets.
    \item We use SkMTEB to evaluate 31 open-weight and proprietary embedding models spanning compact, mid-sized, and large multilingual systems.
    \item We adapt Multilingual E5 into compact Slovak embedding models with vocabulary trimming and targeted fine-tuning, then ablate trimming, fine-tuning, and prompt usage.
    \item We openly release all models, datasets, and code: models and datasets are available at \url{https://huggingface.co/collections/slovak-nlp/skmteb} and the code at \url{https://github.com/slovak-nlp/skmteb}.
\end{itemize}

\begin{figure*}[ht]
    \centering
    \includegraphics[width=\textwidth]{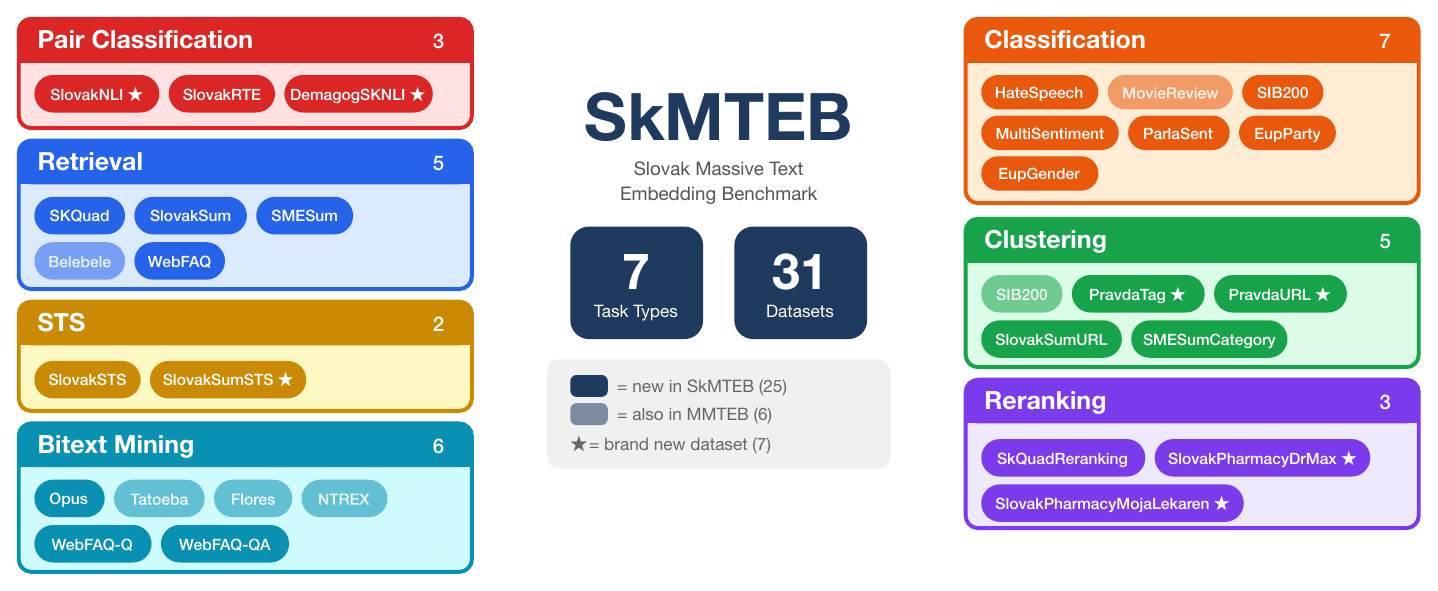}
    \caption{Overview of the SkMTEB benchmark comprising 31 datasets across 7 task types. Lighter shading indicates datasets also present in MMTEB (6), while solid shading marks datasets not in MMTEB (25). Datasets marked with {\small$\bigstar$} are 7 brand-new datasets created specifically for this work.}
    \label{fig:skmteb-overview}
\end{figure*}

\section{Related Work}

Text embeddings have become fundamental components in modern NLP, powering semantic search, retrieval-augmented generation~\cite{lewis-etal-2020-rag}, clustering, and classification systems. The field has witnessed a consistent trend toward larger models: from early approaches averaging Word2Vec~\cite{mikolov2013word2vec} or GloVe~\cite{pennington-etal-2014-glove} word vectors, through BERT-based sentence encoders~\cite{devlin-etal-2019-bert,reimers-gurevych-2019-sentence-bert}, to today's billion-parameter embedding models. Recent state-of-the-art models exemplify this scaling trend---Multilingual E5~\cite{wang-etal-2024-multilingual-e5} ranges from 118M to 560M parameters, BGE-M3~\cite{chen-etal-2024-bge-m3} contains 568M parameters, Jina Embeddings v3~\cite{sturua-etal-2024-jina-v3} reaches 570M parameters, and the latest Qwen3-Embedding models~\cite{qwen3-embedding-2025} scale up to 8B parameters. Alongside this scaling trend, recent releases increasingly emphasise instruction tuning, multi-task training, and practical efficiency: Jina Embeddings v4~\cite{gunther-etal-2025-jina-v4} extends the line toward unified multimodal multilingual retrieval, Nomic Embed v2-MoE~\cite{nussbaum-etal-2025-nomic-moe} applies sparse Mixture-of-Experts to text embeddings, and compact open models such as Granite Embedding~\cite{granite-embedding-2025} and EmbeddingGemma~\cite{embeddinggemma-2025} target deployment-constrained settings. While these large models achieve impressive benchmark performance, their computational requirements pose significant challenges for practical deployment, particularly for applications requiring low latency, edge deployment, or cost-effective scaling.

For under-resourced languages, this scale-efficiency tension is compounded by a structural inefficiency: multilingual models allocate substantial capacity to high-resource languages~\cite{wu-dredze-2020-languages}, and their embedding matrices---often comprising 30\%--40\% of total parameters~\cite{ushio-etal-2023-vocab-trimming}---are dominated by tokens that are irrelevant to any single target language. Several approaches have emerged to address this efficiency gap. Vocabulary Trimming~\cite{ushio-etal-2023-vocab-trimming} removes tokens irrelevant to the target language, reducing model size by up to 66\% while preserving performance, with recent work on Dutch~\cite{banar-etal-2025-mteb-nl} demonstrating its applicability to embedding models. Static embedding approximations like Static-Similarity~\cite{sentence-transformers-2024-static} achieve 100--400$\times$ faster inference while retaining 86\%--95\% of performance, and distillation approaches~\cite{reimers-gurevych-2020-multilingual-distillation} enable creating smaller models from larger teachers---illustrating that practical deployment often requires trading some accuracy for substantial efficiency gains.

Evaluating these trade-offs requires robust benchmarks. The Massive Text Embedding Benchmark (MTEB)~\cite{muennighoff-etal-2023-mteb} established a standardized evaluation framework spanning classification, clustering, semantic textual similarity, retrieval, and pair classification tasks with 56 English datasets. While MTEB primarily targets English, it catalyzed the development of language-specific benchmarks providing comparable depth: C-MTEB~\cite{xiao-etal-2023-cpack} for Chinese (35 datasets), PL-MTEB~\cite{poswiata-etal-2024-plmteb} for Polish (28 datasets), the Scandinavian Embedding Benchmark~\cite{enevoldsen-etal-2024-seb} for Danish, Norwegian, and Swedish, FR-MTEB~\cite{ciancone-etal-2024-frmteb} for French, ruMTEB~\cite{snegirev-etal-2024-rumteb} for Russian (23 datasets), ArabicMTEB~\cite{bhatia-etal-2025-arabicmteb} for Arabic (94 datasets), and FaMTEB~\cite{famteb-2025} for Persian.

The recent MMTEB~\cite{enevoldsen-etal-2025-mmteb} takes a complementary approach, prioritizing breadth across 250+ languages over depth in any single language. This design necessarily yields shallow per-language coverage: Slovak is represented by only 8 tasks in MMTEB---just 14\% of English MTEB's depth and 29\% of PL-MTEB's coverage. These 8 tasks consist primarily of subsets from multilingual datasets (SIB-200, FLORES, Tatoeba) that, while enabling cross-lingual comparison, lack Slovak-specific evaluation scenarios such as native retrieval benchmarks, domain-specific tasks, or temporally grounded evaluation. SkMTEB addresses this gap with 31 datasets---nearly 4$\times$ MMTEB's Slovak coverage---spanning domains (medical, fact-checking, parliamentary), task formulations (summarization-as-retrieval, URL-based clustering), and temporal ranges (2000--2025). Only 6 datasets overlap, ensuring the benchmarks are complementary: MMTEB enables cross-lingual comparison, while SkMTEB provides the depth needed to diagnose Slovak-specific model behavior.

\section{The SkMTEB Benchmark}

SkMTEB comprises 31 datasets across 7 task types following the MTEB framework~\cite{muennighoff-etal-2023-mteb}, providing comprehensive coverage of Slovak text embedding evaluation. The benchmark spans diverse domains including news, government, social media, reviews, and encyclopedic content, with temporal coverage from 2000 to 2025. For each task type below, we first define the task and evaluation metrics, then describe the Slovak datasets we include. Full task and dataset metadata are provided in Appendix~\ref{app:tasks}, and curation details for newly created datasets are summarized in Appendices~\ref{app:slovak-pharmacy-reranking} and~\ref{app:slovak-sts-synthetic}.
Figure~\ref{fig:skmteb-overview} provides a compact overview of the benchmark composition.

\subsection{Retrieval (5 datasets)}

From a large corpus of documents and a set of query texts, the model must retrieve (rank) the most relevant documents per query by embedding similarity. The standard metrics include nDCG@k (primarily nDCG@10)~\cite{Wang2013}, MRR@k, MAP@k, precision@k, and recall@k. One of the first to evaluate the retrieval of Slovak information was \cite{Hladek2016}.

\textbf{SKQuadRetrieval} is a question-answering retrieval task based on the SK-QuAD dataset~\cite{Hladek2023}, evaluating search performance with relevance-scored answers from encyclopedic content. \textbf{SlovakSumRetrieval}~\cite{ondrejova-suppa-2024-slovaksum} and \textbf{SMESumRetrieval}~\cite{suppa-adamec-2020-summarization} reformulate news summarization datasets as retrieval tasks, using article abstracts as queries to retrieve full documents from collections of 200k+ and 80k articles, respectively. \textbf{BelebeleRetrieval}~\cite{bandarkar2023belebele} provides machine reading comprehension across 122 language variants, while \textbf{WebFAQRetrieval}~\cite{dinzinger2025webfaq} contains question-answer pairs from web FAQ pages across 75 languages.

\subsection{Reranking (3 datasets)}

Given a query and a list of candidate reference texts (some relevant, some irrelevant), the model must produce a ranking of the candidates according to relevance to the query (via embedding similarity). The ranked lists are evaluated using ranking metrics: Mean Average Precision (MAP) and Mean Reciprocal Rank (MRR), with MAP as the principal metric.

\textbf{SkQuadReranking} derives from SK-QuAD~\cite{Hladek2023} for article retrieval reranking with manually annotated hard-negatives. \textbf{SlovakPharmacyDrMaxReranking} and \textbf{SlovakPharmacyMojaLekarenReranking} are reranking datasets built from pharmacist Q\&A content; curation details are provided in Appendix~\ref{app:slovak-pharmacy-reranking}.

\subsection{Classification (7 datasets)}

Texts with labels are split into train and test sets; embeddings are computed, and then a simple classifier (e.g., logistic regression) is trained on the training embeddings and evaluated on the test embeddings. The main evaluation metric is classification accuracy (with optional F1 and average precision).

\textbf{SlovakHateSpeechClassification.v2}~\cite{Sokolova2025} annotates social media posts for hateful or offensive language. \textbf{SlovakMovieReviewSentimentClassification.v2}~\cite{vstefanik2023resources} provides binary sentiment classification from 30k+ Slovak movie reviews (2002--2020). \textbf{SIB200Classification}~\cite{adelani2023sib} is the largest public topic classification dataset, covering 205 languages with 7 topics. \textbf{MultilingualSentimentClassification}~\cite{mollanorozy-etal-2023-cross} spans 30 languages with binary sentiment labels. \textbf{SlovakParlaSentClassification}~\cite{mochtak-etal-2024-parlasent} contains 3-level sentiment annotations from parliamentary debates. \textbf{MultiEupSlovakPartyClassification} and \textbf{MultiEupSlovakGenderClassification}~\cite{yang-etal-2024-language-bias} predict political groups and gender from European Parliament speeches (2020--2024).

\subsection{Clustering (5 datasets)}

A collection of texts (sentences, paragraphs) is embedded and then clustered into groups. Clustering quality is assessed with label-agnostic metrics such as $V$-measure, which is insensitive to label permutations.

\textbf{SIB200ClusteringS2S}~\cite{adelani2023sib} clusters up to 1,004 documents into 7 thematic topics. \textbf{PravdaSKTagClustering} and \textbf{PravdaSKURLClustering} cluster Pravda.sk news articles by tags and URL structure into 50 categories. \textbf{SlovakSumURLClustering}~\cite{ondrejova-suppa-2024-slovaksum} and \textbf{SMESumCategoryClustering}~\cite{suppa-adamec-2020-summarization} organize news articles into 12 and 11 editorial categories, respectively.

\subsection{Bitext Mining (6 datasets)}

Given two sets of sentences in different languages, the model must find for each sentence in the first set its best matching translation in the second set using embedding similarity. The performance is primarily measured by F1 (and also precision and recall).

\textbf{OpusSlovakEnglishBitextMining}~\cite{zhang2020improving} provides Slovak-English parallel sentences from OPUS-100 (2000--2020). \textbf{TatoebaBitextMining}~\cite{tatoeba} and \textbf{FloresBitextMining}~\cite{goyal2022flores} offer multilingual parallel corpora with Slovak support. \textbf{NTREXBitextMining}~\cite{federmann-etal-2022-ntrex} provides Slovak translations of a multilingual news corpus. \textbf{WebFAQBitextMiningQuestions} and \textbf{WebFAQBitextMiningQAs}~\cite{dinzinger2025webfaq} enable cross-lingual question and Q\&A pair retrieval across 75 languages. To keep evaluation focused, we restrict each task to Slovak-English and, where available, Slovak-Czech pairs.

\subsection{Pair Classification (3 datasets)}

The task is to take a pair of texts and decide whether they are equivalent (e.g., paraphrase, duplicate) or not. The model embeds both texts and computes a distance or similarity; the thresholding then yields binary predictions, and metrics such as average precision (cosine) are used.

\textbf{SlovakNLI} contains handwritten premise-hypothesis pairs for natural language inference (entailment vs. contradiction). \textbf{SlovakRTE}~\cite{suppa-etal-2025-sklep} is a professionally translated and human-verified recognizing textual entailment dataset from the skLEP benchmark. \textbf{DemagogSKNLI} creates NLI pairs from Demagog.sk fact-checking data (2010--2025), pairing evidence with political statements for claim verification.

\subsection{Semantic Textual Similarity (2 datasets)}

For a pair of sentences, the objective is to predict a continuous similarity score. The embedding similarity (e.g., cosine) is correlated with the human-annotated similarity scores, with Spearman correlation being the main evaluation metric.

\textbf{SlovakSTS}~\cite{suppa-etal-2025-sklep} is a Slovak translation of the GLUE STSb dataset, providing human-verified semantic similarity scores (0--5) for sentence pairs from blogs and news. \textbf{SlovakSumSTS} is a synthetic dataset derived from the SlovakSum news corpus~\cite{ondrejova-suppa-2024-slovaksum}, with LLM-generated similarity scores that have been human-verified; curation details are provided in Appendix~\ref{app:slovak-sts-synthetic}.

\subsection{Baseline Models}

We evaluate a diverse set of embedding models to establish baselines on SkMTEB. Our selection spans several axes: (i) historical anchors that established multilingual embedding benchmarks, (ii) models with explicit Slovak-language support, (iii) coverage of the size–quality trade-off from lightweight to large-scale, (iv) architectural diversity (dense, sparse, MoE, instruction-tuned), and (v) leading proprietary APIs as upper-bound references. We organize the baselines by model family and briefly motivate each group in historical order.

\paragraph{Sentence-Transformers and early multilingual baselines.}
\texttt{LaBSE} arrived as one of the first truly strong language-agnostic sentence encoders and quickly became a default reference for multilingual retrieval and bitext mining, making it an essential historical anchor for SkMTEB.~\cite{feng-etal-2022-labse} We also include the widely used \texttt{paraphrase-multilingual-mpnet-base-v2} and the compact \texttt{paraphrase-multilingual-MiniLM-L12-v2}, which set practical baselines for multilingual similarity and remain common production choices due to their balance of quality and efficiency.~\cite{reimers-gurevych-2019-sentence-bert} The more recent \texttt{static-similarity-mrl-multilingual-v1} represents a different trade-off: an explicitly speed‑oriented model that prioritized ultra-fast CPU inference at the time of release and thus captures the efficiency end of the multilingual spectrum.~\cite{sentence-transformers-2024-static}

\paragraph{Slovak-specific baselines.}
\texttt{slovakbert-sts-stsb} adapts SlovakBERT for sentence similarity, providing a localized baseline that reflects Slovak linguistic idiosyncrasies rather than cross-lingual transfer alone.~\cite{pikuliak-etal-2022-slovakbert} \texttt{slovakbert-skquad-mnlr} extends this idea to retrieval-style supervision, grounding a Slovak-specific model in QA-derived ranking signals and offering a native reference point for retrieval tasks at the time Slovak resources were scarce.~\cite{pikuliak-etal-2022-slovakbert,Hladek2023}

\paragraph{Multilingual E5 family.}
The \texttt{multilingual-e5-small}, \texttt{multilingual-e5-base}, and \texttt{multilingual-e5-large} models established a strong, consistently competitive multilingual baseline that was easy to use and broadly effective across task types, making E5 the default yardstick for many benchmarks when it appeared.~\cite{wang-etal-2024-multilingual-e5} The later \texttt{multilingual-e5-large-instruct} variant brought instruction tuning into this family, reflecting the field’s shift toward promptable embeddings and improved transfer across heterogeneous tasks.~\cite{wang-etal-2024-multilingual-e5}

\paragraph{Modern multilingual retrieval families.}
\texttt{bge-m3} was notable for unifying dense, sparse, and multi-vector retrieval in a single model, signaling a move toward multi-functionality rather than single‑purpose encoders.~\cite{chen-etal-2024-bge-m3} \texttt{gte-multilingual-base} emphasized long-context retrieval and elastic embeddings, capturing the growing demand for longer documents and reranking‑style pipelines at the time of release.~\cite{zhang-etal-2024-mgte} \texttt{snowflake-arctic-embed-l-v2.0} represents Snowflake's second-generation large embedding model, bringing improved multilingual retrieval performance and long-context support through a refined training pipeline that targets practical enterprise retrieval workloads.~\cite{yu-etal-2024-arctic}

\paragraph{Nomic and Jina model lines.}
\texttt{nomic-embed-text-v1.5} popularized long‑context embedding with Matryoshka representations, letting practitioners trade embedding size for speed without retraining and making it a frequent baseline in applied settings.~\cite{nussbaum-etal-2024-nomic-v2} \texttt{nomic-embed-text-v2-moe} advanced this line with sparse Mixture‑of‑Experts, offering a higher quality‑efficiency frontier that reflected broader trends in scalable embedding training.~\cite{nussbaum-etal-2025-nomic-moe} On the Jina side, \texttt{jina-embeddings-v3} introduced task‑specific LoRA adapters and Matryoshka learning, while \texttt{jina-embeddings-v4} expanded to multimodal multilingual retrieval; together they capture a progression from flexible text embeddings to unified text‑image representations that became increasingly important in the field.~\cite{sturua-etal-2024-jina-v3,gunther-etal-2025-jina-v4}

\paragraph{Recent large-scale embedding families.}
The \texttt{granite-embedding-107m-multilingual} and \texttt{granite-embedding-278m-multilingual} models represent IBM's modern embedding line, providing compact and mid‑sized baselines with contemporary training recipes that were competitive at release time.~\cite{granite-embedding-2025} \texttt{Qwen3-Embedding-0.6B} marks the entry of a strong foundation‑model lineage into embeddings, offering an efficient Qwen‑family baseline.~\cite{qwen3-embedding-2025} Finally, \texttt{embeddinggemma-300m} reflects the push toward lightweight, high‑quality open embeddings, making it a timely reference point for practical deployments with limited compute.~\cite{embeddinggemma-2025}

\paragraph{Proprietary API models.}
To complement open-weight baselines, we include leading proprietary embedding services. OpenAI's \texttt{text-embedding-3-small} and \texttt{text-embedding-3-large}\footnote{\url{https://platform.openai.com/docs/guides/embeddings}} represent the current generation of their embedding API, with the large variant offering higher capacity at increased cost. Cohere's \texttt{embed-v4.0}\footnote{\url{https://docs.cohere.com/docs/embed}} provides another commercial reference point with strong multilingual capabilities. Amazon's \texttt{titan-embed-text-v2}\footnote{\url{https://docs.aws.amazon.com/bedrock/latest/userguide/titan-embedding-models.html}} completes the proprietary baselines as AWS's current-generation embedding API with strong multilingual capabilities. These API models establish upper bounds for what commercial solutions offer on Slovak text, though their closed nature limits reproducibility and architectural analysis.

All models are evaluated using their default configurations and recommended preprocessing steps to ensure fair comparison across different architectures and training approaches.

\section{Adapting Embedding Models for Slovak}

Beyond benchmarking, we explore training Slovak-specific embedding models by fine-tuning SlovakBERT~\cite{pikuliak-etal-2022-slovakbert} and models from the Multilingual E5 family~\cite{wang-etal-2024-multilingual-e5}.

\paragraph{Training Data}
We use datasets from the skLEP benchmark~\cite{suppa-etal-2025-sklep}: SK-SQuAD~\cite{Hladek2023} (72K query-context pairs), NLI translated from XNLI~\cite{conneau-etal-2018-xnli} (393K pairs), STS from GLUE STS-B~\cite{cer-etal-2017-semeval} (6K pairs), and RTE from GLUE~\cite{wang-etal-2018-glue} (2.5K pairs). We also experiment with Slovak Web QA (967K pairs from WebFAQ~\cite{dinzinger2025webfaq} and MFAQ~\cite{de-bruyn-etal-2021-mfaq}), where hard negatives are randomly sampled answers from the same domain; we later exclude this dataset as this construction does not consistently provide meaningful contrastive signal.

\paragraph{Training Configuration}
We use mean pooling with a max sequence length of 256, multi-task learning with Cosine Similarity Loss for STS and Multiple Negatives Ranking Loss~\cite{henderson2017mnrl} for other tasks. Training uses a batch size of 32, a learning rate $2 \times 10^{-5}$ with a linear warmup (10\% of training steps), and 3 epochs. All experiments use a single NVIDIA H100 GPU with random seed 42. Training completes in under 1 hour per model variant. Full hyperparameters are provided in Appendix~\ref{app:reproducibility}.

\paragraph{Vocabulary Trimming}
We apply Vocabulary Trimming (VT)~\cite{ushio-etal-2023-vocab-trimming} to create compact Slovak-specific models from multilingual E5 encoders, following recent work on Dutch embeddings~\cite{banar-etal-2025-mteb-nl}. VT removes vocabulary tokens irrelevant to the target language; we retain 60K tokens (from 250K) following the recommendation of \citet{ushio-etal-2023-vocab-trimming}, who found this threshold balances vocabulary coverage with model efficiency across multiple target languages. Token retention is determined by frequency in FineWeb2-Slovak,\footnote{\url{https://huggingface.co/datasets/ivykopal/fineweb2-slovak}} a quality-filtered Slovak web corpus~\cite{penedo2024fineweb2}. We apply VT before fine-tuning (Pre-FT VT), reducing both model size and training time. This yields size reductions of 62\% for E5-small (118M $\rightarrow$ 45M) and 35\% for E5-large (560M $\rightarrow$ 365M).

\begin{table*}[ht]
\centering
\small
\resizebox{\textwidth}{!}{%
\begin{tabular}{lrrrrrrrrr}
\toprule
& \multicolumn{2}{c}{\textbf{Average Across}} & \multicolumn{7}{c}{\textbf{Average per Task Type}} \\
\cmidrule(lr){2-3} \cmidrule(lr){4-10} \\
\textbf{Model} (\(\downarrow\)) & \textbf{All} & \textbf{Type} & \textbf{Btxt} & \textbf{Clf} & \textbf{Clust} & \textbf{PrClf} & \textbf{Rrnk} & \textbf{Rtrvl} & \textbf{STS} \\
\midrule 

\textcolor{gray!75}{Number of datasets (\(\rightarrow\))} & \textcolor{gray!75}{(31)} & \textcolor{gray!75}{(7)} & \textcolor{gray!75}{(6)} & \textcolor{gray!75}{(7)} & \textcolor{gray!75}{(5)} & \textcolor{gray!75}{(3)} & \textcolor{gray!75}{(3)} & \textcolor{gray!75}{(5)} & \textcolor{gray!75}{(2)} \\
\midrule 

\midrule
\multicolumn{10}{c}{\textit{Small models (<130M)}} \\
\rowcolor{gray!15} \texttt{e5-sk-small} \textsubscript{\textcolor{gray!60}{\tiny \textbf{(45M)}}} & 70.56 & 72.01 & 91.34 & 60.84 & 40.95 & 66.05 & 84.94 & 78.64 & 81.32 \\
\texttt{granite-embedding-107m-multilingual} \textsubscript{\textcolor{gray!60}{\tiny \textbf{(107M)}}} & 65.81 & 67.31 & 85.85 & 56.04 & 40.80 & 60.88 & 78.10 & 70.93 & 78.54 \\
\texttt{static-similarity-mrl-multilingual-v1} \textsubscript{\textcolor{gray!60}{\tiny \textbf{(108M)}}} & 58.51 & 60.90 & 87.98 & 48.87 & 17.47 & 63.66 & 70.76 & 59.35 & 78.22 \\
\texttt{multilingual-e5-small} \textsubscript{\textcolor{gray!60}{\tiny \textbf{(118M)}}} & 70.32 & 71.78 & 91.29 & 59.81 & 41.09 & 64.66 & 85.13 & 79.21 & 81.25 \\
\texttt{paraphrase-multilingual-MiniLM-L12-v2} \textsubscript{\textcolor{gray!60}{\tiny \textbf{(118M)}}} & 67.86 & 69.16 & 94.91 & 60.83 & 37.44 & 67.82 & 73.46 & 66.03 & 83.62 \\
\texttt{slovakbert-skquad-mnlr} \textsubscript{\textcolor{gray!60}{\tiny \textbf{(125M)}}} & 67.49 & 68.91 & 85.07 & 62.49 & 38.07 & 65.13 & 76.91 & 72.82 & 81.90 \\
\texttt{slovakbert-sts-stsb} \textsubscript{\textcolor{gray!60}{\tiny \textbf{(125M)}}} & 63.44 & 65.37 & 81.75 & 63.05 & 31.82 & \underline{68.90} & 67.15 & 59.20 & 85.69 \\
\rowcolor{gray!15} \texttt{sturovec-base} \textsubscript{\textcolor{gray!60}{\tiny \textbf{(125M)}}} & 68.99 & 70.13 & 86.97 & 62.19 & 41.55 & 63.27 & 80.66 & 76.54 & 79.73 \\
\midrule
\multicolumn{10}{c}{\textit{Base models (>=130M, <350M)}} \\
\texttt{nomic-embed-text-v1.5} \textsubscript{\textcolor{gray!60}{\tiny \textbf{(137M)}}} & 51.52 & 55.03 & 46.64 & 51.40 & 28.77 & 61.98 & 68.60 & 55.39 & 72.43 \\
\texttt{granite-embedding-278m-multilingual} \textsubscript{\textcolor{gray!60}{\tiny \textbf{(278M)}}} & 67.69 & 69.09 & 87.56 & 57.81 & 42.41 & 61.40 & 81.25 & 73.92 & 79.30 \\
\texttt{multilingual-e5-base} \textsubscript{\textcolor{gray!60}{\tiny \textbf{(278M)}}} & 72.39 & 73.57 & 94.76 & 62.35 & 40.91 & 64.10 & 85.71 & 83.57 & 83.59 \\
\texttt{paraphrase-multilingual-mpnet-base-v2} \textsubscript{\textcolor{gray!60}{\tiny \textbf{(278M)}}} & 70.34 & 71.67 & 96.23 & 64.05 & 38.95 & 68.21 & 76.79 & 70.03 & 87.44 \\
\texttt{gte-multilingual-base} \textsubscript{\textcolor{gray!60}{\tiny \textbf{(305M)}}} & 71.76 & 73.26 & 94.36 & 61.70 & 39.28 & 67.24 & 83.04 & 81.63 & 85.57 \\
\texttt{embeddinggemma-300m} \textsubscript{\textcolor{gray!60}{\tiny \textbf{(308M)}}} & 69.25 & 70.56 & 83.89 & 62.08 & 43.76 & 65.54 & 81.00 & 78.40 & 79.28 \\
\texttt{nomic-embed-text-v2-moe} \textsubscript{\textcolor{gray!60}{\tiny \textbf{(330M)}}} & 72.58 & 73.84 & 90.03 & 63.98 & 43.53 & 64.19 & 85.15 & 85.37 & 84.60 \\
\midrule
\multicolumn{10}{c}{\textit{Large models (>=350M)}} \\
\rowcolor{gray!15} \texttt{e5-sk-large} \textsubscript{\textcolor{gray!60}{\tiny \textbf{(365M)}}} & 74.70 & 75.88 & 96.39 & 66.34 & 41.43 & 67.32 & 87.81 & 85.60 & 86.25 \\
\texttt{LaBSE} \textsubscript{\textcolor{gray!60}{\tiny \textbf{(471M)}}} & 66.44 & 67.52 & \textbf{97.48} & 58.73 & 34.28 & 64.39 & 73.52 & 63.39 & 80.85 \\
\texttt{multilingual-e5-large} \textsubscript{\textcolor{gray!60}{\tiny \textbf{(560M)}}} & 74.25 & 75.49 & 96.29 & 65.34 & 40.35 & 66.78 & \underline{87.96} & 85.80 & 85.90 \\
\texttt{multilingual-e5-large-instruct} \textsubscript{\textcolor{gray!60}{\tiny \textbf{(560M)}}} & \textbf{77.49} & \textbf{78.44} & \underline{97.09} & \underline{70.28} & \textbf{49.69} & \textbf{70.55} & 86.49 & 86.08 & \underline{88.86} \\
\texttt{bge-m3} \textsubscript{\textcolor{gray!60}{\tiny \textbf{(568M)}}} & 74.43 & 75.55 & 96.29 & 66.81 & 39.97 & 67.03 & 86.72 & 85.63 & 86.39 \\
\texttt{snowflake-arctic-embed-l-v2.0} \textsubscript{\textcolor{gray!60}{\tiny \textbf{(568M)}}} & 72.54 & 73.63 & 93.46 & 63.40 & 40.41 & 63.11 & 87.11 & 85.17 & 82.76 \\
\texttt{jina-embeddings-v3} \textsubscript{\textcolor{gray!60}{\tiny \textbf{(572M)}}} & 75.10 & 76.20 & 96.43 & 66.89 & 44.63 & 66.44 & 83.44 & 85.77 & \textbf{89.82} \\
\texttt{Qwen3-Embedding-0.6B} \textsubscript{\textcolor{gray!60}{\tiny \textbf{(596M)}}} & 70.53 & 71.80 & 90.12 & 62.84 & 44.28 & 64.46 & 82.02 & 75.73 & 83.14 \\
\texttt{jina-embeddings-v4} \textsubscript{\textcolor{gray!60}{\tiny \textbf{(3.8B)}}} & 72.44 & 73.87 & 90.49 & 62.99 & 44.65 & 64.43 & 85.25 & 83.64 & 85.65 \\
\texttt{Qwen3-Embedding-4B} \textsubscript{\textcolor{gray!60}{\tiny \textbf{(4B)}}} & 73.70 & 74.96 & 94.21 & 64.26 & 44.22 & 65.82 & 86.29 & 84.11 & 85.82 \\
\texttt{Qwen3-Embedding-8B} \textsubscript{\textcolor{gray!60}{\tiny \textbf{(8B)}}} & 74.53 & 75.75 & 94.43 & 65.94 & 43.36 & 66.65 & 87.04 & \underline{86.27} & 86.54 \\
\midrule
\multicolumn{10}{c}{\textit{API access models}} \\
\texttt{embed-v4.0} & 71.26 & 72.82 & 90.23 & 60.12 & 40.13 & 63.12 & 87.45 & 85.71 & 82.95 \\
\texttt{text-embedding-3-small} & 70.48 & 71.39 & 91.82 & 62.13 & 43.64 & 63.24 & 82.37 & 76.96 & 79.56 \\
\texttt{text-embedding-3-large} & 75.07 & 75.89 & 96.79 & 66.91 & 44.22 & 66.58 & 86.96 & 85.55 & 84.21 \\
\texttt{gemini-embedding-001} & \underline{77.23} & \underline{78.07} & 96.76 & \textbf{71.01} & \underline{46.26} & 67.08 & \textbf{88.27} & \textbf{88.34} & 88.74 \\
\texttt{amazon-titan-embed-text-v2} & 67.24 & 69.21 & 84.98 & 59.26 & 33.39 & 62.78 & 83.99 & 77.43 & 82.67 \\
\bottomrule
\end{tabular}
}
\caption{SkMTEB results summary (percent). The table reports the average performance across all tasks (\textbf{All}) and the unweighted average across task types (\textbf{Type}), followed by per-task-type averages for Bitext Mining (Btxt), Classification (Clf), Clustering (Clust), Pair Classification (PrClf), Reranking (Rrnk), Retrieval (Rtrvl), and STS. The best result per task is \textbf{bolded} with the runner-up \underline{underlined}.}
\label{tab:mteb_results}
\end{table*}

\subsection{Fine-tuned Models}

We explore two approaches to training Slovak embedding models.

\paragraph{SlovakBERT with Full Training Data}
Our initial approach fine-tunes SlovakBERT~\cite{pikuliak-etal-2022-slovakbert} on the complete training dataset, including the Slovak Web QA triplets. The resulting model, \texttt{sturovec-base}, achieves reasonable performance but falls short of the unfine-tuned \texttt{multilingual-e5-small} baseline (68.99 vs.\ 70.32 average score), despite using 1.4M training examples. Upon analyzing the Slovak Web QA triplets, we identified quality issues stemming from the automated hard negative sampling process, where random answers from the same domain do not consistently provide meaningful contrastive signal. This motivated a refined approach for subsequent experiments.

\paragraph{Vocabulary-Trimmed E5 Models}
Based on these observations, we fine-tune models from the Multilingual E5 family~\cite{wang-etal-2024-multilingual-e5} \emph{without} the Slovak Web QA triplets, using only the higher-quality skLEP datasets (SK-SQuAD, NLI, STS, RTE). Prior to fine-tuning, we apply vocabulary trimming to reduce the models to 60K tokens based on FineWeb2-Slovak frequency statistics. This substantially reduces model size: \texttt{e5-sk-small} contains 45M parameters compared to 118M in the original \texttt{multilingual-e5-small}---a reduction of over 60\%. For the large variant, \texttt{e5-sk-large} is reduced from 560M to 365M parameters.

Both models are fine-tuned for 3 epochs using the same hyperparameters as the SlovakBERT experiments. Following standard E5 practice, we prepend \texttt{query:} and \texttt{passage:} prefixes to queries and documents, respectively, during both training and inference.

\section{SkMTEB Results}

Our main results are summarized in Table~\ref{tab:mteb_results}, which reports average scores by task type, and in Table~\ref{tab:mteb_results_clf}, which details per-task classification performance. Figure~\ref{fig:mteb_results_per_size} and Figure~\ref{fig:skmteb-pareto} visualize the relationship between model size and average performance.

Among evaluated models, \texttt{multilingual-e5-large-instruct} achieves the highest overall score (77.49), followed closely by \texttt{gemini-embedding-001} (77.23). These instruction-tuned and large-scale models excel particularly in classification and clustering tasks, where they outperform smaller alternatives by significant margins. The Multilingual E5 family demonstrates strong performance across model sizes, with even the small variant (118M) achieving competitive results. Very large models do not consistently outperform 500M--600M alternatives on SkMTEB: \texttt{jina-embeddings-v4} (3.8B, 72.44) trails \texttt{snowflake-arctic-embed-l-v2.0} (568M, 72.54) and \texttt{nomic-embed-text-v2-moe} (330M, 72.58), and only narrowly edges \texttt{multilingual-e5-base} (278M, 72.39). This suggests diminishing returns from scale alone for Slovak embedding tasks.

Task difficulty varies substantially across the benchmark. Bitext mining proves largely solved, with most models reaching F1 above 90---reflecting the relative ease of cross-lingual alignment for Slovak-English and Slovak-Czech pairs. In contrast, clustering remains challenging, with V-measure ranging from 17 to 50, indicating significant room for improvement. Reranking and retrieval tasks show strong performance from E5-family models, while STS benefits from models with explicit similarity training objectives like \texttt{jina-embeddings-v3} (89.82). Existing Slovak-specific models trained primarily for NLU tasks (\texttt{slovakbert-skquad-mnlr}, \texttt{slovakbert-sts-stsb}) underperform compared to multilingual alternatives, highlighting the need for dedicated embedding model development.

Our vocabulary-trimmed E5 models demonstrate competitive performance, matching proprietary API models: \texttt{e5-sk-small} (70.56) performs on par with \texttt{text-embedding-3-small} (70.48), while \texttt{e5-sk-large} (74.70) achieves comparable results to \texttt{text-embedding-3-large} (75.07). TOST equivalence testing \citep{lakens2017equivalence} confirms practical equivalence: 90\% CIs for both comparisons fall within $\pm$2 points. Crucially, our models offer practical advantages: they are open-weight, can run locally without API costs, and their reduced size enables higher throughput---making them accessible for practitioners working with Slovak text.

\section{Ablation Study}

We conduct ablation experiments to isolate the contributions of vocabulary trimming (VT), fine-tuning (FT), and inference prompts to our final models. Table~\ref{tab:ablation} summarizes results for the E5-small and E5-large model families.

\begin{table}[t]
\centering
\small
\resizebox{\columnwidth}{!}{%
\begin{tabular}{lccrrl}
\toprule
\textbf{Model Variant} & \textbf{VT} & \textbf{FT} & \textbf{Size} & \textbf{Avg} & \textbf{$\Delta$} \\
\midrule
\texttt{mE5-small} (baseline) & & & 118M & 70.32 & --- \\
\quad + VT & \checkmark & & 45M & 70.45 & \textcolor{teal}{+0.13} \\
\quad + FT & & \checkmark & 118M & 70.58 & \textcolor{teal}{+0.26} \\
\quad + VT + FT & \checkmark & \checkmark & 45M & 70.56 & \textcolor{teal}{+0.24} \\
\quad + VT + FT + prompt & \checkmark & \checkmark & 45M & \textbf{71.07} & \textcolor{teal}{+0.75} \\
\midrule
\texttt{mE5-large} (baseline) & & & 560M & 74.25 & --- \\
\quad + VT & \checkmark & & 365M & 74.56 & \textcolor{teal}{+0.31} \\
\quad + FT & & \checkmark & 560M & 74.46 & \textcolor{teal}{+0.21} \\
\quad + VT + FT & \checkmark & \checkmark & 365M & 74.70 & \textcolor{teal}{+0.45} \\
\quad + VT + FT + prompt & \checkmark & \checkmark & 365M & \textbf{74.72} & \textcolor{teal}{+0.47} \\
\bottomrule
\end{tabular}%
}
\caption{Ablation study on vocabulary trimming (VT), fine-tuning (FT), and prompt usage. The $\Delta$ column shows change relative to the baseline. Size reduction from VT: 62\% for small, 35\% for large.}
\label{tab:ablation}
\end{table}

\paragraph{Vocabulary Trimming}
VT reduces model size dramatically---from 118M to 45M parameters for E5-small (62\% reduction) and from 560M to 365M for E5-large (35\% reduction)---while preserving or slightly improving performance (+0.13 for small, +0.31 for large).

\paragraph{Fine-tuning}
Fine-tuning on skLEP data (excluding the noisy Slovak Web QA triplets) provides modest but consistent improvements: +0.26 for E5-small and +0.21 for E5-large.

\paragraph{Combined Effects}
Combining VT and FT yields additive benefits for the large model (+0.45 total improvement over baseline while reducing size by 35\%). For the small model, the combined approach achieves nearly the same performance as FT alone but with 62\% fewer parameters.

\paragraph{Inference Prompts}
Adding \texttt{query:}/\texttt{passage:} prefixes during both training and inference provides additional improvements: +0.51 for E5-small (70.56 → 71.07) and +0.02 for E5-large (74.70 → 74.72). The larger effect on the small model suggests that explicit query-passage distinction particularly benefits models with limited capacity.

\paragraph{Cross-lingual Transfer Preservation}
A natural concern with aggressive vocabulary trimming is that removing non-Slovak tokens could degrade cross-lingual transfer. Table~\ref{tab:crosslingual} reports per-pair F1 on the six cross-lingual bitext mining tasks in SkMTEB for the original and trimmed E5 models. Differences are small in both directions (maximum absolute change 0.92~F1 for the small model and 0.14~F1 for the large model), and several pairs marginally improve after trimming. Targeted vocabulary reduction thus preserves Slovak-English and Slovak-Czech transfer, addressing a practical concern for deployments that process code-mixed or adjacent-language content.

\begin{table}[t!]
\centering
\small
\begin{tabular}{lrrrr}
\toprule
 & \multicolumn{2}{c}{\textbf{E5-small}} & \multicolumn{2}{c}{\textbf{E5-large}} \\
\cmidrule(lr){2-3} \cmidrule(lr){4-5}
\textbf{Task (pair)} & \textbf{Orig.} & \textbf{Trim.} & \textbf{Orig.} & \textbf{Trim.} \\
\midrule
Flores (ces-slk)  & 99.87 & 99.74 & 99.87 & 99.87 \\
Flores (eng-slk)  & 95.58 & 95.72 & 100.00 & 100.00 \\
NTREX (ces-slk)   & 98.08 & 98.11 & 99.22 & 99.27 \\
NTREX (eng-slk)   & 94.64 & 95.56 & 98.80 & 98.66 \\
Tatoeba (slk-eng) & 82.69 & 82.56 & 93.22 & 93.22 \\
Opus (slk-eng)    & 77.08 & 76.95 & 86.21 & 86.27 \\
\bottomrule
\end{tabular}
\caption{Cross-lingual bitext mining F1 for original vs.\ vocabulary-trimmed Multilingual E5 models. Differences stay within 1~F1 point on all pairs, averaging 0.25~F1 for the small model and 0.04~F1 for the large model.}
\label{tab:crosslingual}
\end{table}

\section{Conclusion}

We presented SkMTEB, the first comprehensive text embedding benchmark for Slovak, comprising 31 datasets across 7 task types. Our evaluation reveals that large instruction-tuned multilingual models achieve strong cross-lingual transfer to Slovak, while compact alternatives like multilingual-e5-small offer competitive performance suitable for practical deployment. Existing Slovak-specific NLU models do not transfer well to embedding tasks, highlighting the need for dedicated development.

We also show that combining vocabulary trimming with fine-tuning on curated Slovak data yields compact, competitive embedding models with modest computational resources. Our 45M parameter model achieves 91\% of the performance of the best 560M parameter multilingual model, demonstrating that vocabulary trimming combined with targeted fine-tuning offers a practical path to efficient, language-specific embeddings. Ablations show that trimming reduces E5-small by 62\% and E5-large by 35\% while preserving Slovak-English and Slovak-Czech bitext mining within 1~F1 point. We release SkMTEB, our vocabulary-trimmed models, and all associated code under open-source licenses.

Beyond Slovak, four findings should transfer to other under-resourced languages: (1) NLU-tuned monolingual encoders underperform on embedding tasks, making language-specific embedding evaluation necessary even where NLU benchmarks exist; (2) vocabulary trimming generalizes from NLU to embedding models, yielding 35--62\% parameter reductions with negligible in-language or cross-lingual degradation; (3) 4B--8B-parameter embedding models rarely outperform their 500M--600M counterparts on a single target language, suggesting embedding-specific scaling trends distinct from those observed for generative LLMs; and (4) the full SkMTEB pipeline---benchmark construction, broad evaluation, trimming, and fine-tuning---can be replicated with modest compute (under one GPU-hour per adapted model), providing a practical template for teams working on similar languages.

\section*{Limitations}

\paragraph{Benchmark coverage.} While SkMTEB substantially expands Slovak embedding evaluation from 8 to 31 tasks, gaps remain. Several datasets rely on machine translation from English sources with native speaker post-editing, which may not fully capture Slovak-specific linguistic phenomena or cultural contexts. Domain coverage skews toward news, Wikipedia, and web content; specialized domains such as legal, medical, or technical Slovak are underrepresented.

\paragraph{Translated vs.\ native data.} A portion of our benchmark derives from translated datasets (NLI, STS, RTE). While we employed machine translation with native speaker post-editing, translated data may exhibit translationese artifacts and fail to reflect authentic Slovak language use. Future work should prioritize natively authored Slovak datasets.

\paragraph{Model selection.} Our baseline evaluation, while extensive, cannot cover all available embedding models. We focused on models with documented multilingual support, including both open-weight models and select proprietary APIs; however, very recent releases may not be represented. Additionally, computational constraints limited evaluation of the largest models (8B+ parameters) across all tasks.

\paragraph{Temporal snapshot.} Both the benchmark and model evaluations represent a snapshot in time. The embedding model landscape evolves rapidly, and newer models may substantially outperform those evaluated here. We encourage ongoing community contributions to keep evaluations current.

\paragraph{Vocabulary trimming trade-offs.} While vocabulary trimming reduces model size, it may affect performance on code-mixed or multilingual Slovak text. Our per-pair analysis (Table~\ref{tab:crosslingual}) shows that Slovak-English and Slovak-Czech transfer is preserved within 1~F1 point on all tested pairs, but we do not evaluate performance on non-Slavic, non-English pairs or on sentences that interleave multiple languages.

\section*{Ethics Statement}

\paragraph{Data licensing.} All datasets included in SkMTEB are released under permissive licenses or with explicit permission for research use. We document licensing information for each dataset in Appendix~\ref{app:tasks} and release our benchmark under a license compatible with downstream research and development.

\paragraph{Privacy and offensive content.} The datasets in SkMTEB are derived from publicly available sources (Wikipedia, news articles, web content, parliamentary proceedings) or existing research datasets. We did not collect new data from human subjects. For newly curated datasets (pharmacy Q\&A, news clustering), we verified that source data does not contain personally identifying information beyond public figures mentioned in news contexts. The hate speech classification dataset contains offensive language necessary for the task; we include content warnings in dataset documentation and recommend appropriate handling during model development. Parliamentary and news datasets may contain politically sensitive content reflecting public discourse.

\paragraph{Potential biases.} Embedding models and the datasets used to train and evaluate them may encode societal biases present in their source data. SkMTEB inherits biases from its constituent datasets, and strong benchmark performance does not guarantee fair or unbiased model behavior. We encourage users to conduct bias audits appropriate to their deployment contexts.

\paragraph{Intended use.} SkMTEB is intended for research evaluation of text embedding models for Slovak. While we hope it enables the development of better Slovak NLP systems---for example in public-sector search, accessibility, and moderation over Slovak text---users should validate model performance on their specific use cases rather than relying solely on benchmark scores. Retrieval and clustering over Slovak political discourse could also enable surveillance or political profiling; we recommend context-appropriate governance when deploying models trained on or evaluated with politically salient datasets such as \texttt{DemagogSKNLI} and \texttt{SlovakParlaSentClassification}.

\paragraph{Environmental impact.} Running the full SkMTEB evaluation suite requires substantial computation. We report approximate compute requirements in the appendix to enable carbon footprint estimation. Our work on vocabulary trimming and efficient adaptation aims to reduce the computational burden of deploying embedding models.

\paragraph{Use of AI assistants.} In addition to GPT-5 for dataset generation (Appendix~\ref{app:slovak-sts-synthetic}), we used AI assistants (Claude, ChatGPT) to aid with figure visualization, code refactoring, and English grammar and vocabulary checking, as the authors are non-native English speakers. All AI-generated content was reviewed and verified by the authors.

\section*{Acknowledgments}

This study was funded by the Ministry of Education, Research, Development and Youth of the Slovak Republic under the project KEGA 049TUKE-4/2024, VEGA 1/0685/26 and by the Slovak Research and Development Agency under the project APVV-22-0414.

This work was partially funded by European Union, under the project lorAI - Low Resource Artificial Intelligence, GA No. 101136646, https://doi.org/10.3030/101136646. It was also partially funded by the EU NextGenerationEU through the Recovery and Resilience Plan for Slovakia under the project No. 09I02-03-V01-00029.

Part of the research results was obtained using the computational resources procured in the
national project National competence centre for high performance computing (project code:
311070AKF2) funded by European Regional Development Fund, EU Structural Funds Informatization
of society, Operational Program Integrated Infrastructure.

\bibliography{custom}
\newpage
\appendix

\section{SkMTEB Pareto Frontier}
\label{app:skmteb-pareto}

\begin{figure}[ht]
    \centering
    \includegraphics[width=\columnwidth]{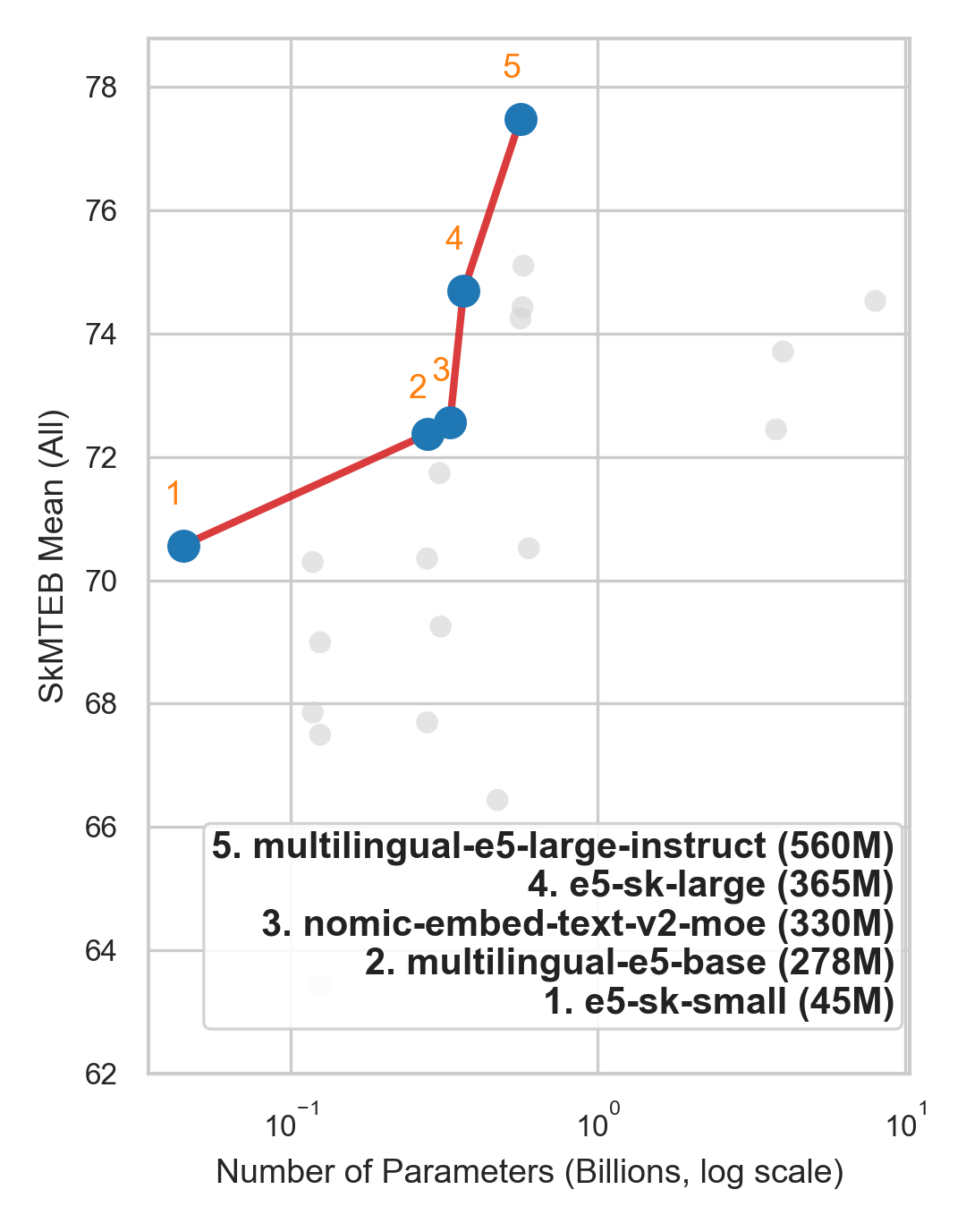}
    \caption{SkMTEB Pareto frontier (All). Each point is a model with mean SkMTEB score (y-axis) versus parameter count in billions (x-axis, log scale). Gray points show all evaluated models; the red polyline connects the Pareto-optimal set (no model simultaneously smaller and higher-scoring). Blue points are the frontier models; orange numbers above each point correspond to the ordered list in the legend (bottom-right), which reports model names and sizes. The y-axis is truncated to start at 62 to emphasize differences among competitive models.}
    \label{fig:skmteb-pareto}
\end{figure}

\section{Slovak Pharmacy Reranking (Dataset Curation)}
\label{app:slovak-pharmacy-reranking}
Data were scraped from two Slovak pharmacy websites, MojaLekaren and DrMax. 
Each of them has a forum page where customers can ask health-related questions. A certified pharmacist will answer the question.
After scraping, the data were cleaned, formatted and anonymized by removing personal names. Duplicate questions were removed. Each dataset includes customer questions as reranking queries and corresponding pharmacist answers as the only positive answer in the reranking dataset. Therefore, a question-answer pair is a query-positive pair. 

In addition, the DrMax dataset tags were also scraped. These tags provide a description of the pair. The most common tags include \textit{interactions}, \textit{pain}, \textit{side effects}, and \textit{pregnancy}. 
The MojaLekaren dataset contains both tags and categories, which were merged into a single set of tags. Labels \textit{eyes}, \textit{hair}, and \textit{pain} were included in the tags and categories. The most common tags were \textit{advice of a pharmacist}, \textit{digestion}, and \textit{skin}.

For each query-positive pair, four negative answers are assigned. Negative answers are responses to different questions. 
The process of assigning negatives was as follows. If the query-positive pair has more than four tags, the four most frequent tags were selected. If exactly four tags are present, all of them are used. For each tag, a negative answer is retrieved from a different question-answer pair that shares the same tag. 

If the tag is unique, a random answer has been selected. Each selected negative answer is then compared with the original positive answer using cosine similarity, and only answers whose similarity falls within a threshold are retained. If the similarity constraint is not satisfied, a different answer is selected. Sentence embeddings are computed using the paraphrase-multilingual-MiniLM-L12-v2 model, and cosine similarity is calculated using the scikit-learn library.

If fewer than four tags are available for a given query-answer pair, the following procedure is applied. For each available tag, one negative answer is selected as described above. 
The remaining negative answers are selected from answers that share the most frequent tag. If no suitable answers are found for that tag, the next most frequent tag is considered.
If no suitable answers are available, a random answer is selected.

The dataset mojalekaren-reranking contains 738 rows, with a lower similarity threshold of 0.55 and an upper threshold of 0.85. Dataset drmax-reranking contains 4676 rows, with a lower similarity threshold of 0.3 and an upper threshold of 0.9. 
The upper threshold was selected after manually reviewing several examples to determine whether a negative sample qualifies as a hard negative or a false negative. 
The lower threshold was established to filter out overly simplistic negatives, thereby ensuring the inclusion of sufficiently challenging samples for the model.

\begin{figure*}[t]
\centering
\includegraphics[width=1\linewidth]{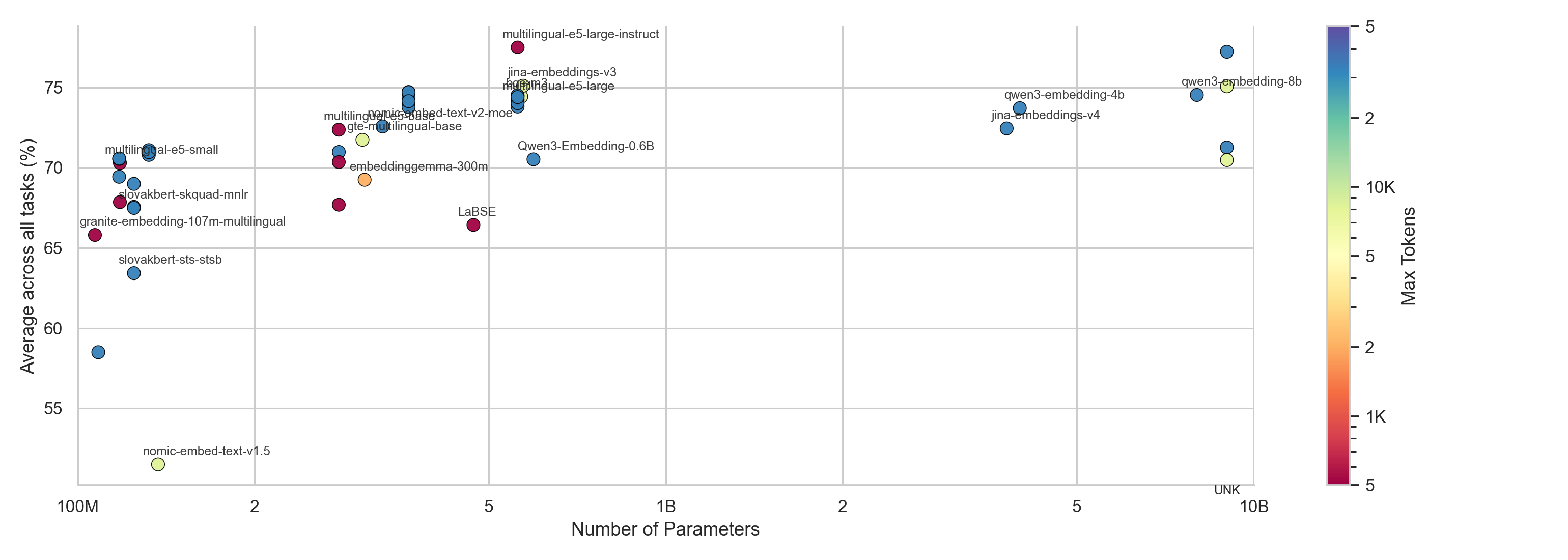}
\caption{Model Size vs. Average Performance}
\label{fig:mteb_results_per_size}
\end{figure*}

\section{Slovak STS Synthetic (Dataset Curation)}
\label{app:slovak-sts-synthetic}

To construct the dataset, we paired original sentences from the Slovak Summarization Dataset \cite{ondrejova-suppa-2024-slovaksum} with synthetic counterparts generated by the GPT-5 model.
For every original sentence, GPT-5 was prompted to produce six distinct variations, each corresponding to one of the six semantic similarity levels (0--5) defined by \cite{agirre2013sem}.
The STS score definitions used for generation are summarized in Table~\ref{tab:sts_score}.

\begin{table}[h!]
\centering
\begin{tabular}{c p{5cm}}  
\toprule
\textbf{STS Score} & \textbf{Description} \\
\midrule
0 & The two sentences are on different topics. \\
1 & The two sentences are not equivalent, but are on the same topic. \\
2 & The two sentences are not equivalent, but share some details. \\
3 & The two sentences are roughly equivalent, but some important information differs or is missing. \\
4 & The two sentences are mostly equivalent, but some unimportant details differ. \\
5 & The two sentences are completely equivalent, as they mean the same thing. \\
\bottomrule
\end{tabular}
\caption{Semantic Textual Similarity (STS) score definitions~\cite{agirre2013sem}}
\label{tab:sts_score}
\end{table}

The following constraints were applied to the original sentences: they had to contain at least 60 characters and no more than 200 characters, and they had to consist of a single sentence (multi-sentence examples were omitted). GPT-5 was required to generate exactly six sentences, one for each similarity score. 
The dataset generation process was repeated four times, with 20 sentences per iteration, with the prompt refined at each stage to address any generation issues (e.g., similarity between score 4 and 5 or score 1 and 2). GPT-5 was provided with two examples from the authors in the prompt, including explanations for why each sentence received its score and a detailed description of all similarity scores. The temperature parameter was set to 1.
A total of 6594 sentence pairs were generated.

The annotation process took place after generation.
A random sample of 300 sentence pairs was selected from the full generated dataset, with 50 pairs drawn from each similarity score. The annotation guidelines did not explicitly specify the target distribution of similarity scores.
Three native Slovak speakers served as annotators: two are coauthors of this work, and the third is a colleague who volunteered without any expectation of remuneration. All annotators received information about the task (evaluating each pair based on the similarity score), explanations for each STS score, and five examples with explanations.
The first two examples were identical to those used in the GPT-5 prompt. 
The remaining three examples were pairs of original and generated sentences, together with an explanation of why GPT-5 generated that particular sentence for that similarity score. These sentences were selected from the entire STS dataset but were not part of the annotated subset.
All three annotators independently annotated the same set of 300 sentence pairs.

After annotation, the results were validated. Pairwise inter-annotator agreement scores, including Cohen's kappa, are reported in Table~\ref{tab:pairwise_agreement}. All Cohen's kappa values exceed 0.5, indicating moderate agreement. Across all annotators, Fleiss' kappa is 0.56 and Krippendorff's alpha is 0.92.

During the annotation process, the annotators found a Hungarian sentence within the original text, although the corresponding generated sentence was in Slovak. 
After this, the authors manually reviewed the entire dataset to verify language consistency. Two Hungarian sentences were found, and 12 pairs were deleted from the dataset. The dataset was subsequently cleaned and formatted. 
To prevent data contamination, all original sentences selected for annotation and all their generated sentences were removed from the non-annotated split.
This reduced the number of original sentences from 1099 to 821. 

\begin{table*}[ht]
\centering
\begin{tabular}{lcccc p{5cm}}
\toprule
\textbf{Annotator pair} & \textbf{Overall Accuracy} & \textbf{Pearson Corr.} & \textbf{Spearman Corr.} & \textbf{Cohen's $\kappa$} \\
\midrule
{[1,2]} & 0.67 & 0.94 & 0.94 & 0.60 \\
{[1,3]} & 0.63 & 0.91 & 0.92 & 0.56 \\
{[2,3]} & 0.61 & 0.93 & 0.93 & 0.53 \\
\bottomrule
\end{tabular}
\caption{Pairwise agreement between annotators}
\label{tab:pairwise_agreement}
\end{table*}

The final dataset is divided into two parts: a train split consisting of the non-annotated portion (4926 rows), where the similarity score ranges from 0 to 5, and a test split consisting of the annotated portion (298 rows), where the similarity score is the average score of three annotators.

\subsection{Generation Prompt}

The full system prompt used for generating sentence variants is provided below. The prompt is written entirely in Slovak to ensure natural, native-quality outputs. It instructs the model to:

\begin{enumerate}
    \item Read an original Slovak sentence from the SlovakSum corpus.
    \item Generate six variant sentences, one for each STS score (0--5).
    \item Provide a Slovak explanation justifying each score assignment.
    \item Follow detailed score boundary definitions distinguishing between adjacent scores (e.g., score 4 vs.\ 5, score 1 vs.\ 2).
    \item Maintain natural Slovak language quality suitable for native speakers.
    \item Return output in a structured JSON format.
\end{enumerate}

The prompt (in Figure \ref{fig:sts-prompt}) includes two complete worked examples demonstrating expected outputs for different source sentences. Key design choices include:
\begin{itemize}
    \item \textbf{Explicit boundary definitions:} The prompt carefully distinguishes adjacent scores to reduce annotator confusion (e.g., score 4 requires ``nearly equivalent with minor generalizations'' while score 5 requires ``complete paraphrase with all information preserved'').
    \item \textbf{Anti-repetition constraints:} For score 0 (different topic), the prompt explicitly prohibits defaulting to common topics like weather or tourism.
    \item \textbf{Quality control:} The prompt instructs the model to internally verify each sentence sounds natural to a native Slovak speaker before outputting.
\end{itemize}

\begin{figure*}[p]
\begin{promptbox}[System Prompt for STS Generation (Slovak) --- Abbreviated]
Si expert na slovenský jazyk, parafrázovanie a anotáciu sémantickej podobnosti.

\textbf{Tvoja úloha:}\\
1. Prečítať si jednu slovenskú vetu (ORIGINÁL).\\
2. Pre KAŽDÉ skóre podobnosti 0--5 (STS) vytvor:\\
\hspace*{3mm}-- jednu slovenskú vetu (VARIANT),\\
\hspace*{3mm}-- jedno krátke vysvetlenie v slovenčine.

\tcblower

\textbf{DEFINÍCIE SKÓRE 0--5}\\[2pt]
\textit{Všeobecné pravidlo:} „Dôležitá informácia" = hlavná udalosť, aktér, miesto, čas, množstvo, základná pointa. Ak zmeníš alebo vynecháš dôležitú informáciu, skóre NEMÔŽE byť 4 ani 5.\\[4pt]
\textbf{Skóre 0} -- ÚPLNE INÁ TÉMA\\
Vety sú o úplne odlišných veciach. Žiadne zmysluplné tematické prepojenie.\\[2pt]
\textbf{Skóre 1} -- ROVNAKÁ ŠIROKÁ TÉMA, ALE TAKMER ŽIADNE DETAILY SPOLOČNÉ\\
Vety majú spoločnú tému, hovoria však o iných konkrétnych udalostiach.\\[2pt]
\textbf{Skóre 2} -- NIE SÚ EKVIVALENTNÉ, ALE ZDIEĽAJÚ NIEKTORÉ DETAILY\\
Vety sa týkajú podobnej témy a zdieľajú niektoré konkrétne prvky.\\[2pt]
\textbf{Skóre 3} -- PRIBLIŽNÁ EKVIVALENCIA, ALE DÔLEŽITÁ INFORMÁCIA SA LÍŠI\\
Hlavná udalosť je podobná, ALE zmení sa aspoň JEDEN DÔLEŽITÝ aspekt.\\[2pt]
\textbf{Skóre 4} -- TAKMER EKVIVALENTNÉ, LÍŠIA SA LEN MENEJ PODSTATNÉ DETAILY\\
Hlavná udalosť, aktéri, miesto, čas a čísla sú rovnaké. Rozdiely sú minimálne.\\[2pt]
\textbf{Skóre 5} -- ÚPLNE ROVNAKÝ VÝZNAM (PARAFRÁZA)\\
Všetky dôležité informácie sú zachované. VARIANT je parafráza s inými slovami.\\[4pt]
\textbf{VÝSTUPNÝ FORMÁT} -- JSON objekt:\\
\{"0": \{"sentence": "...", "explanation": "..."\}, ..., "5": \{...\}\}
\end{promptbox}
\caption{Abbreviated system prompt for STS sentence generation (Slovak). The full prompt includes detailed score boundary definitions, two complete worked examples, and additional quality guidelines; see supplementary materials for the complete version.}
\label{fig:sts-prompt}
\end{figure*}

\onecolumn

\section{Model Parameter Size and Embedding Dimension}

\begin{table*}[t]
\centering
\small
\begin{tabular}{lrr}
\toprule
\textbf{Model} (\(\downarrow\)) & \textbf{Params} & \textbf{Dim} \\
\midrule
\multicolumn{3}{c}{\textit{Small models (<130M)}} \\
\rowcolor{gray!15} e5-sk-small & 45M & 384 \\
granite-embedding-107m-multilingual & 107M & 384 \\
static-similarity-mrl-multilingual-v1 & 108M & 1024 \\
multilingual-e5-small & 118M & 384 \\
paraphrase-multilingual-MiniLM-L12-v2 & 118M & 768 \\
slovakbert-skquad-mnlr & 125M & 768 \\
slovakbert-sts-stsb & 125M & 768 \\
\rowcolor{gray!15} sturovec-base & 125M & 768 \\
\midrule
\multicolumn{3}{c}{\textit{Base models (>=130M, <350M)}} \\
nomic-embed-text-v1.5 & 137M & 768 \\
granite-embedding-278m-multilingual & 278M & 768 \\
multilingual-e5-base & 278M & 768 \\
paraphrase-multilingual-mpnet-base-v2 & 278M & 768 \\
gte-multilingual-base & 305M & 768 \\
embeddinggemma-300m & 308M & 768 \\
nomic-embed-text-v2-moe & 330M & 1024 \\
\midrule
\multicolumn{3}{c}{\textit{Large models (>=350M)}} \\
\rowcolor{gray!15} e5-sk-large & 365M & 1024 \\
LaBSE & 471M & 768 \\
multilingual-e5-large & 560M & 1024 \\
multilingual-e5-large-instruct & 560M & 1024 \\
bge-m3 & 568M & 1024 \\
snowflake-arctic-embed-l-v2.0 & 568M & 1024 \\
jina-embeddings-v3 & 572M & 1024 \\
Qwen3-Embedding-0.6B & 596M & 1024 \\
jina-embeddings-v4 & 3.8B & 2048 \\
Qwen3-Embedding-4B & 4B & 2560 \\
Qwen3-Embedding-8B & 8B & 4096 \\
\midrule
\multicolumn{3}{c}{\textit{API access models}} \\
embed-v4.0 & - & 1536 \\
text-embedding-3-small & - & 1536 \\
text-embedding-3-large & - & 3072 \\
gemini-embedding-001 & - & 3072 \\
amazon-titan-embed-text-v2 & - & 1024 \\
\bottomrule
\end{tabular}
\caption{Model size and embedding dimension. Models are grouped by parameter-size buckets with API-access models listed last. This table is intended as a companion reference to the main results tables and uses the same model ordering, naming, and highlighting conventions.}
\label{tab:mteb_results_model_param_dim}
\end{table*}

\newpage

\section{SkMTEB Task Catalogue}
\label{app:tasks}

This appendix provides comprehensive descriptions of all tasks included in the SkMTEB benchmark, organized by task type. The benchmark comprises 31 distinct task definitions across 7 categories, with several bitext mining tasks evaluated on multiple language pair subsets. For each task, we briefly describe the data source, creation methodology, temporal coverage, licensing, and evaluation specifics.

\paragraph{B6: Statistics for data.} We report dataset sizes and evaluation split details where applicable in the task catalogue below (e.g., evaluation size, temporal coverage, and annotations). For datasets we create or curate, Appendix~\ref{app:slovak-pharmacy-reranking} and Appendix~\ref{app:slovak-sts-synthetic} provide additional statistics such as row counts, sampling constraints, and annotation sizes.

\input{datasheet_table}

\subsection{Retrieval Tasks}

\subsubsection{SKQuadRetrieval}
\textbf{Source:} SK-QuAD dataset~\cite{Hladek2023} \\
\textbf{HuggingFace:} \texttt{TUKE-KEMT/retrieval-skquad} \\
\textbf{Description:} A question-answering retrieval task that evaluates Slovak search performance using questions and answers derived from the SK-QuAD dataset. The task measures relevance with scores assigned to answers based on their relevancy to corresponding questions. \\
\textbf{Domain:} Encyclopaedic \\
\textbf{Task Subtype:} Question answering \\
\textbf{Annotations:} Human-annotated relevance scores \\
\textbf{License:} CC-BY-NC-SA-4.0 \\
\textbf{Main Metric:} nDCG@10

\subsubsection{SlovakSumRetrieval}
\textbf{Source:} SlovakSum dataset~\cite{ondrejova-suppa-2024-slovaksum} \\
\textbf{HuggingFace:} \texttt{NaiveNeuron/slovaksum} \\
\textbf{Description:} A Slovak news summarization dataset consisting of over 200,000 news articles with titles and short abstracts obtained from multiple Slovak newspapers. Originally intended as a summarization task, reformulated to a retrieval task where article abstracts serve as queries to retrieve full documents. \\
\textbf{Domain:} News, Social, Web \\
\textbf{Task Subtype:} Article retrieval \\
\textbf{Temporal Coverage:} 2015--2022 \\
\textbf{Annotations:} Derived from document structure \\
\textbf{License:} OpenRAIL \\
\textbf{Main Metric:} nDCG@10 \\
\textbf{Evaluation Size:} 600 query-document pairs

\subsubsection{SMESumRetrieval}
\textbf{Source:} SMESum dataset~\cite{suppa-adamec-2020-summarization} \\
\textbf{HuggingFace:} \texttt{NaiveNeuron/SMESum} \\
\textbf{Description:} A Slovak news summarization dataset consisting of 80,000 news articles with titles and introductions from the SME news portal. Reformulated as a retrieval task where article introductions serve as queries to retrieve full documents. \\
\textbf{Domain:} News, Social, Web \\
\textbf{Task Subtype:} Article retrieval \\
\textbf{Temporal Coverage:} 2013--2019 \\
\textbf{Annotations:} Derived from document structure \\
\textbf{License:} Not specified \\
\textbf{Main Metric:} nDCG@10 \\
\textbf{Evaluation Size:} 600 query-document pairs

\subsubsection{BelebeleRetrieval}
\textbf{Source:} Belebele benchmark~\cite{bandarkar2023belebele} \\
\textbf{HuggingFace:} \texttt{facebook/belebele} \\
\textbf{Description:} A multiple-choice machine reading comprehension dataset spanning 122 language variants. For Slovak (slk\_Latn), the task involves retrieving relevant passages given questions. \\
\textbf{Domain:} Web, News \\
\textbf{Task Subtype:} Question answering \\
\textbf{Annotations:} Expert-annotated \\
\textbf{License:} CC-BY-SA-4.0 \\
\textbf{Main Metric:} nDCG@10

\subsubsection{WebFAQRetrieval}
\textbf{Source:} WebFAQ corpus~\cite{dinzinger2025webfaq} \\
\textbf{HuggingFace:} \texttt{mteb/WebFAQRetrieval} \\
\textbf{Description:} A broad-coverage corpus of natural question-answer pairs gathered from FAQ pages on the web, covering 75 languages including Slovak. \\
\textbf{Domain:} Web \\
\textbf{Task Subtype:} Question answering \\
\textbf{Temporal Coverage:} 2022--2024 \\
\textbf{Annotations:} Derived from FAQ structure \\
\textbf{License:} CC-BY-4.0 \\
\textbf{Main Metric:} nDCG@10

\subsection{Semantic Textual Similarity Tasks}

\subsubsection{SlovakSTS}
\textbf{Source:} skLEP benchmark~\cite{suppa-etal-2025-sklep} \\
\textbf{HuggingFace:} \texttt{slovak-nlp/sklep} (subset: sts) \\
\textbf{Description:} Professional Slovak translation of the original GLUE STSb dataset. Contains sentence pairs with human-annotated similarity scores ranging from 0 (completely dissimilar) to 5 (completely similar). \\
\textbf{Domain:} Blog, News \\
\textbf{Task Subtype:} Textual Entailment \\
\textbf{Temporal Coverage:} 2025 \\
\textbf{Annotations:} Human-annotated, machine-translated and verified \\
\textbf{License:} CC-BY-SA-4.0 \\
\textbf{Main Metric:} Spearman correlation

\subsubsection{SlovakSumSTS}
\textbf{Source:} Synthetic dataset from SlovakSum \\
\textbf{HuggingFace:} \texttt{slovak-nlp/slovak-sts-synthetic} \\
\textbf{Description:} Sentence pairs for semantic textual similarity scoring in Slovak. Pairs were generated using text from the SlovakSum dataset, where an LLM created corresponding sentence pairs for each STS score (0--5). The test split pairs were verified by human annotators. \\
\textbf{Domain:} News \\
\textbf{Task Subtype:} Textual Entailment \\
\textbf{Temporal Coverage:} 2025 \\
\textbf{Annotations:} LM-generated and human-reviewed \\
\textbf{License:} CC-BY-NC-4.0 \\
\textbf{Main Metric:} Spearman correlation

\subsection{Pair Classification Tasks}

\subsubsection{SlovakNLI}
\textbf{Source:} Handwritten Slovak NLI dataset \\
\textbf{HuggingFace:} \texttt{natalia-nk/NLI-SK-annotated} \\
\textbf{Description:} Slovak handwritten annotated Natural Language Inference dataset containing premise-hypothesis pairs. Labels indicate entailment or contradiction relationships. \\
\textbf{Domain:} News, Web \\
\textbf{Task Subtype:} Textual Entailment \\
\textbf{Temporal Coverage:} 2024--2025 \\
\textbf{Annotations:} Human-annotated \\
\textbf{License:} Not specified \\
\textbf{Main Metric:} Average Precision (max\_ap)

\subsubsection{SlovakRTE}
\textbf{Source:} skLEP benchmark~\cite{suppa-etal-2025-sklep} \\
\textbf{HuggingFace:} \texttt{slovak-nlp/sklep} (subset: rte) \\
\textbf{Description:} Slovak Recognizing Textual Entailment dataset. Professional translation and human verification of English RTE datasets for Slovak. Binary classification task (entailment vs. not entailment). \\
\textbf{Domain:} News, Web \\
\textbf{Task Subtype:} Textual Entailment \\
\textbf{Temporal Coverage:} 2025 \\
\textbf{Annotations:} Human-annotated, machine-translated and verified \\
\textbf{License:} CC-BY-SA-4.0 \\
\textbf{Main Metric:} Average Precision (max\_ap)

\subsubsection{DemagogSKNLI}
\textbf{Source:} Demagog.sk fact-checking portal \\
\textbf{HuggingFace:} \texttt{NaiveNeuron/DemagogSK} \\
\textbf{Description:} Slovak Natural Language Inference dataset created from Demagog.sk fact-checking data. Evidence-claim pairs where professional fact-checkers' analysis (evidence) is paired with political statements (claims). Labels indicate whether evidence supports (Pravda) or refutes (Nepravda) the claim. \\
\textbf{Domain:} Government, News \\
\textbf{Task Subtype:} Claim verification \\
\textbf{Temporal Coverage:} 2010--2025 \\
\textbf{Annotations:} Expert-annotated by fact-checkers \\
\textbf{License:} Not specified \\
\textbf{Main Metric:} Average Precision (max\_ap)

\subsection{Classification Tasks}

\subsubsection{SlovakHateSpeechClassification.v2}
\textbf{Source:} TUKE-KEMT hate speech dataset \\
\textbf{HuggingFace:} \texttt{mteb/slovak\_hate\_speech} \\
\textbf{Description:} Social network posts with human annotations for hateful or offensive language in Slovak. Binary classification (toxic vs. not toxic). Version 2 corrects errors from the original dataset. \\
\textbf{Domain:} Social \\
\textbf{Task Subtype:} Sentiment/Hate speech \\
\textbf{Temporal Coverage:} 2024 \\
\textbf{Annotations:} Human-annotated \\
\textbf{License:} CC-BY-SA-4.0 \\
\textbf{Main Metric:} Accuracy

\subsubsection{SlovakMovieReviewSentimentClassification.v2}
\textbf{Source:} CSFD movie database~\cite{vstefanik2023resources} \\
\textbf{HuggingFace:} \texttt{mteb/slovak\_movie\_review\_sentiment} \\
\textbf{Description:} User reviews of movies from the CSFD movie database with binary sentiment classes (positive, negative). Version 2 corrects errors from the original dataset. \\
\textbf{Domain:} Reviews \\
\textbf{Task Subtype:} Sentiment/Hate speech \\
\textbf{Temporal Coverage:} 2002--2020 \\
\textbf{Annotations:} Derived from user ratings \\
\textbf{License:} CC-BY-NC-SA-4.0 \\
\textbf{Main Metric:} Accuracy

\subsubsection{SIB200Classification}
\textbf{Source:} SIB-200 dataset~\cite{adelani2023sib} \\
\textbf{HuggingFace:} \texttt{mteb/sib200} (subset: slk\_Latn) \\
\textbf{Description:} The largest publicly available topic classification dataset based on Flores-200, covering 205 languages and dialects. Annotated for 7 topics: science/technology, travel, politics, sports, health, entertainment, and geography. Labels transferred from English via human translation. \\
\textbf{Domain:} News \\
\textbf{Task Subtype:} Topic classification \\
\textbf{Temporal Coverage:} 2023--2024 \\
\textbf{Annotations:} Expert-annotated for English, human-translated \\
\textbf{License:} CC-BY-SA-4.0 \\
\textbf{Main Metric:} Accuracy

\subsubsection{MultilingualSentimentClassification}
\textbf{Source:} Cross-lingual sentiment dataset~\cite{mollanorozy-etal-2023-cross} \\
\textbf{HuggingFace:} \texttt{mteb/multilingual-sentiment-classification} (subset: slk) \\
\textbf{Description:} Sentiment classification dataset with binary labels (positive vs. negative) covering 30 languages and dialects including Slovak. \\
\textbf{Domain:} Reviews \\
\textbf{Task Subtype:} Sentiment/Hate speech \\
\textbf{Temporal Coverage:} 2022 \\
\textbf{Annotations:} Derived \\
\textbf{License:} Not specified \\
\textbf{Main Metric:} Accuracy

\subsubsection{SlovakParlaSentClassification}
\textbf{Source:} ParlaSent corpus~\cite{mochtak-etal-2024-parlasent} \\
\textbf{HuggingFace:} \texttt{classla/ParlaSent} (subset: SK) \\
\textbf{Description:} Slovak parliamentary sentiment classification from the ParlaSent corpus. Contains sentences from parliamentary debates with 3-level sentiment annotations (negative, neutral, positive). \\
\textbf{Domain:} Government, Spoken \\
\textbf{Task Subtype:} Sentiment/Hate speech \\
\textbf{Temporal Coverage:} 2018 \\
\textbf{Annotations:} Human-annotated \\
\textbf{License:} CC-BY-SA-4.0 \\
\textbf{Main Metric:} Accuracy

\subsubsection{MultiEupSlovakPartyClassification}
\textbf{Source:} Multi-EuP v2 corpus~\cite{yang-etal-2024-language-bias} \\
\textbf{HuggingFace:} \texttt{unimelb-nlp/MultiEup-v2} \\
\textbf{Description:} Multi-class classification to predict European Parliament political group from native Slovak speeches. Uses only speeches originally delivered in Slovak from the Multi-EuP v2 corpus. \\
\textbf{Domain:} Government, Spoken \\
\textbf{Task Subtype:} Topic classification \\
\textbf{Temporal Coverage:} 2020--2024 \\
\textbf{Annotations:} Derived from parliamentary metadata \\
\textbf{License:} CC-BY-4.0 \\
\textbf{Main Metric:} Accuracy

\subsubsection{MultiEupSlovakGenderClassification}
\textbf{Source:} Multi-EuP v2 corpus~\cite{yang-etal-2024-language-bias} \\
\textbf{HuggingFace:} \texttt{unimelb-nlp/MultiEup-v2} \\
\textbf{Description:} Binary classification to predict gender of Members of the European Parliament from native Slovak speeches. Uses only speeches originally delivered in Slovak. \\
\textbf{Domain:} Government, Spoken \\
\textbf{Task Subtype:} Topic classification \\
\textbf{Temporal Coverage:} 2020--2024 \\
\textbf{Annotations:} Derived from parliamentary metadata \\
\textbf{License:} CC-BY-4.0 \\
\textbf{Main Metric:} Accuracy

\subsection{Reranking Tasks}

\subsubsection{SkQuadReranking}
\textbf{Source:} SK-QuAD dataset~\cite{Hladek2023} \\
\textbf{HuggingFace:} \texttt{TUKE-KEMT/reranking-skquad} \\
\textbf{Description:} Article retrieval reranking task derived from SK-QuAD. Given a query and candidate documents, the model must rank documents by relevance. \\
\textbf{Domain:} Encyclopaedic \\
\textbf{Task Subtype:} Article retrieval \\
\textbf{Annotations:} Derived \\
\textbf{License:} CC-BY-SA-4.0 \\
\textbf{Main Metric:} MAP@1000

\subsubsection{SlovakPharmacyDrMaxReranking}
\textbf{Source:} DrMax pharmacy website \\
\textbf{HuggingFace:} \texttt{slovak-nlp/slovak-pharmacy-drmax-reranking} \\
\textbf{Description:} Reranking dataset from Q\&A content on the DrMax pharmacy website. Questions about medications, health conditions, and pharmaceutical advice with answers from qualified pharmacists. \\
\textbf{Domain:} Medical, Web \\
\textbf{Task Subtype:} Article retrieval \\
\textbf{Temporal Coverage:} 2025 \\
\textbf{Annotations:} Derived \\
\textbf{License:} CC-BY-NC-ND-4.0 \\
\textbf{Main Metric:} MAP@1000

\subsubsection{SlovakPharmacyMojaLekarenReranking}
\textbf{Source:} MojaLekaren pharmacy website \\
\textbf{HuggingFace:} \texttt{slovak-nlp/slovak-pharmacy-mojalekaren-reranking} \\
\textbf{Description:} Reranking dataset from Q\&A content on the MojaLekaren pharmacy website. Questions about medications, health conditions, and pharmaceutical advice with answers from qualified pharmacists. \\
\textbf{Domain:} Medical, Web \\
\textbf{Task Subtype:} Article retrieval \\
\textbf{Temporal Coverage:} 2025 \\
\textbf{Annotations:} Derived \\
\textbf{License:} CC-BY-NC-ND-4.0 \\
\textbf{Main Metric:} MAP@1000

\subsection{Clustering Tasks}

\subsubsection{SIB200ClusteringS2S}
\textbf{Source:} SIB-200 dataset~\cite{adelani2023sib} \\
\textbf{HuggingFace:} \texttt{mteb/sib200} (subset: slk\_Latn) \\
\textbf{Description:} Clustering variant of the SIB-200 topic classification dataset. Up to 1,004 documents clustered into 7 thematic topics covering science/technology, travel, politics, sports, health, entertainment, and geography. \\
\textbf{Domain:} News \\
\textbf{Task Subtype:} Thematic clustering \\
\textbf{Annotations:} Expert-annotated for English, human-translated \\
\textbf{License:} CC-BY-SA-4.0 \\
\textbf{Main Metric:} V-measure

\subsubsection{PravdaSKTagClustering}
\textbf{Source:} Pravda.sk news portal \\
\textbf{HuggingFace:} \texttt{NaiveNeuron/pravda-sk-tag-clustering} \\
\textbf{Description:} Clustering of Slovak news articles from Pravda.sk based on article tags. Articles grouped into 50 thematic categories including Slovak politics, international affairs, events, and various topics. Uses title + summary as input. \\
\textbf{Domain:} News \\
\textbf{Task Subtype:} Thematic clustering, Topic classification \\
\textbf{Temporal Coverage:} 2014--2024 \\
\textbf{Annotations:} Derived from article tags \\
\textbf{License:} Not specified \\
\textbf{Main Metric:} V-measure \\
\textbf{Maximum Documents:} 2,048

\subsubsection{PravdaSKURLClustering}
\textbf{Source:} Pravda.sk news portal \\
\textbf{HuggingFace:} \texttt{NaiveNeuron/pravda-sk-url-clustering} \\
\textbf{Description:} Clustering of Slovak news articles from Pravda.sk based on URL structure. Articles organized into 50 editorial categories reflecting the portal's content organization, including news, sports, culture, economy, health, travel, celebrity, and science sections. \\
\textbf{Domain:} News \\
\textbf{Task Subtype:} Thematic clustering, Topic classification \\
\textbf{Temporal Coverage:} 2014--2024 \\
\textbf{Annotations:} Derived from URL structure \\
\textbf{License:} Not specified \\
\textbf{Main Metric:} V-measure \\

\subsubsection{SlovakSumURLClustering}
\textbf{Source:} SlovakSum dataset~\cite{ondrejova-suppa-2024-slovaksum} \\
\textbf{HuggingFace:} \texttt{kiviki/slovaksum-url-clustering} \\
\textbf{Description:} Clustering of Slovak news articles from SlovakSum based on URL structure. Articles organized into 12 editorial categories including sports, culture, economy, health, travel, politics, and technology. \\
\textbf{Domain:} News \\
\textbf{Task Subtype:} Thematic clustering, Topic classification \\
\textbf{Temporal Coverage:} 2015--2022 \\
\textbf{Annotations:} Derived from URL structure \\
\textbf{License:} Not specified \\
\textbf{Main Metric:} V-measure

\subsubsection{SMESumCategoryClustering}
\textbf{Source:} SMESum dataset~\cite{suppa-adamec-2020-summarization} \\
\textbf{HuggingFace:} \texttt{NaiveNeuron/SMESum} \\
\textbf{Description:} Clustering of Slovak news articles from SMESum based on news categories. Articles organized into 11 thematic categories covering politics, economy, sports, culture, and other news domains. Articles with ``none'' category are excluded. \\
\textbf{Domain:} News \\
\textbf{Task Subtype:} Thematic clustering, Topic classification \\
\textbf{Temporal Coverage:} 2013--2019 \\
\textbf{Annotations:} Derived from category metadata \\
\textbf{License:} Not specified \\
\textbf{Main Metric:} V-measure

\subsection{Bitext Mining Tasks}

\subsubsection{OpusSlovakEnglishBitextMining}
\textbf{Source:} OPUS-100 corpus~\cite{zhang2020improving} \\
\textbf{HuggingFace:} \texttt{Helsinki-NLP/opus-100} (subset: en-sk) \\
\textbf{Description:} Slovak-English parallel sentences from OPUS-100, a multilingual dataset with 100 languages for evaluating massively multilingual neural machine translation. \\
\textbf{Domain:} Web, Subtitles, Fiction, Non-fiction \\
\textbf{Temporal Coverage:} 2000--2020 \\
\textbf{Annotations:} Derived from parallel corpora \\
\textbf{License:} Not specified \\
\textbf{Main Metric:} F1

\subsubsection{TatoebaBitextMining}
\textbf{Source:} Tatoeba corpus \\
\textbf{HuggingFace:} \texttt{mteb/tatoeba-bitext-mining} (subset: slk-eng) \\
\textbf{Description:} 1,000 English-aligned sentence pairs for Slovak based on the Tatoeba corpus, a community-contributed collection of sentences and translations. \\
\textbf{Domain:} Written \\
\textbf{Temporal Coverage:} 2006--2021 \\
\textbf{Annotations:} Human-annotated by community contributors \\
\textbf{License:} CC-BY-2.0 \\
\textbf{Main Metric:} F1

\subsubsection{FloresBitextMining}
\textbf{Source:} FLORES benchmark~\cite{goyal2022flores} \\
\textbf{HuggingFace:} \texttt{mteb/FloresBitextMining} \\
\textbf{Subsets:} eng\_Latn-slk\_Latn, ces\_Latn-slk\_Latn \\
\textbf{Description:} Benchmark dataset for machine translation between English and low-resource languages. Slovak pairs include English-Slovak and Czech-Slovak alignments. \\
\textbf{Domain:} Non-fiction, Encyclopaedic \\
\textbf{Temporal Coverage:} 2022 \\
\textbf{Annotations:} Human-annotated \\
\textbf{License:} CC-BY-SA-4.0 \\
\textbf{Main Metric:} F1

\subsubsection{NTREXBitextMining}
\textbf{Source:} NTREX-128 dataset~\cite{federmann-etal-2022-ntrex} \\
\textbf{HuggingFace:} \texttt{mteb/NTREXBitextMining} \\
\textbf{Subsets:} eng\_Latn-slk\_Latn, ces\_Latn-slk\_Latn \\
\textbf{Description:} News Test References for MT Evaluation covering 128 languages. Slovak pairs include English-Slovak and Czech-Slovak alignments with 1,997 parallel sentences each. \\
\textbf{Domain:} News \\
\textbf{Temporal Coverage:} 2019--2022 \\
\textbf{Annotations:} Expert-annotated, human-translated \\
\textbf{License:} CC-BY-SA-4.0 \\
\textbf{Main Metric:} F1

\subsubsection{WebFAQBitextMiningQuestions}
\textbf{Source:} WebFAQ corpus~\cite{dinzinger2025webfaq} \\
\textbf{HuggingFace:} \texttt{PaDaS-Lab/webfaq-bitexts} \\
\textbf{Subsets:} eng-slk, ces-slk \\
\textbf{Description:} Natural FAQ-style question-answer pairs aligned across languages. This task uses questions from aligned Q\&A pairs for cross-lingual retrieval. \\
\textbf{Domain:} Web \\
\textbf{Temporal Coverage:} 2022--2024 \\
\textbf{Annotations:} Human-annotated, human-translated \\
\textbf{License:} CC-BY-4.0 \\
\textbf{Main Metric:} F1

\subsubsection{WebFAQBitextMiningQAs}
\textbf{Source:} WebFAQ corpus~\cite{dinzinger2025webfaq} \\
\textbf{HuggingFace:} \texttt{PaDaS-Lab/webfaq-bitexts} \\
\textbf{Subsets:} eng-slk, ces-slk \\
\textbf{Description:} Natural FAQ-style question-answer pairs aligned across languages. This task uses concatenated question-answer pairs for cross-lingual retrieval. \\
\textbf{Domain:} Web \\
\textbf{Temporal Coverage:} 2022--2024 \\
\textbf{Annotations:} Human-annotated, human-translated \\
\textbf{License:} CC-BY-4.0 \\
\textbf{Main Metric:} F1

\newpage

\section{Reproducibility Details}
\label{app:reproducibility}

To ensure reproducibility, we provide complete training and evaluation details below.

\subsection{Training Hyperparameters}

Table~\ref{tab:hyperparams} summarizes all hyperparameters used for fine-tuning our Slovak embedding models.

\begin{table}[h]
\centering
\small
\begin{tabular}{ll}
\toprule
\textbf{Hyperparameter} & \textbf{Value} \\
\midrule
Base models & mE5-small, mE5-large \\
Pooling strategy & Mean pooling \\
Max sequence length & 256 tokens \\
Batch size & 32 \\
Learning rate & $2 \times 10^{-5}$ \\
LR scheduler & Linear with warmup \\
Warmup ratio & 10\% of steps \\
Training epochs & 3 \\
Optimizer & AdamW \\
Weight decay & 0.01 \\
Random seed & 42 \\
Precision & FP32 \\
\midrule
\multicolumn{2}{c}{\textit{Loss Functions}} \\
STS task & Cosine Similarity Loss \\
Other tasks & Multiple Negatives Ranking Loss \\
\bottomrule
\end{tabular}
\caption{Training hyperparameters for e5-sk models.}
\label{tab:hyperparams}
\end{table}

\subsection{Computational Resources}

All training experiments were conducted on a single NVIDIA H100 GPU (80GB). Training time per model:
\begin{itemize}
    \item \texttt{e5-sk-small} (45M params): $\sim$30 minutes
    \item \texttt{e5-sk-large} (365M params): $\sim$50 minutes
    \item \texttt{sturovec-base} (125M params): $\sim$45 minutes
\end{itemize}

Benchmark evaluation of all 25+ models required approximately 48 GPU-hours on NVIDIA H100.

\subsection{Vocabulary Trimming Procedure}

We apply Pre-FT Vocabulary Trimming following \citet{ushio-etal-2023-vocab-trimming}:
\begin{enumerate}
    \item Compute token frequencies on FineWeb2-Slovak corpus
    \item Rank tokens by frequency in target language
    \item Retain top 60K tokens (recommended threshold from original paper)
    \item Resize embedding and output projection matrices
    \item Fine-tune on downstream tasks
\end{enumerate}

The 60K threshold was chosen based on the original vocabulary trimming paper's recommendation, which found this value provides good coverage while maximizing compression. For comparison, MTEB-NL~\cite{banar-etal-2025-mteb-nl} uses 50K tokens for Dutch.

\subsection{Model and Data Availability}

All released models and datasets are collected on Hugging Face at \url{https://huggingface.co/collections/slovak-nlp/skmteb}, and the code is available at \url{https://github.com/slovak-nlp/skmteb}.

\paragraph{Getting started.} SkMTEB is registered as \texttt{MTEB(slk, v1)} in the \texttt{mteb} library; any embedding model can be evaluated end-to-end with a single command:
\begin{verbatim}
mteb run -m <model-id> -b "MTEB(slk, v1)"
\end{verbatim}

\paragraph{Maintenance and versioning.} Because SkMTEB lives inside the \texttt{mteb} framework, the benchmark remains runnable as the framework evolves. Each task is pinned to a specific Hugging Face dataset revision hash in its definition, so evaluation scores are reproducible across time. We plan to maintain a public leaderboard for community submissions on Hugging Face.

\subsection{Evaluation Protocol}

We use the MTEB evaluation framework~\cite{muennighoff-etal-2023-mteb} with default settings. For classification tasks, we train a logistic regression classifier on embeddings with default scikit-learn parameters. All results are single-run evaluations with seed 42.

\newpage

\section{SkMTEB Classification Results}

\begin{table*}[h]
\centering
\small
\resizebox{\textwidth}{!}{%
\begin{tabular}{lrrrrrrrr}
\toprule
\textbf{Model} (\(\downarrow\)) & \rotatebox{90}{\textbf{MultiEupSlovakGenderClassification}} & \rotatebox{90}{\textbf{MultiEupSlovakPartyClassification}} & \rotatebox{90}{\textbf{MultilingualSentimentClassification}} & \rotatebox{90}{\textbf{SIB200Classification}} & \rotatebox{90}{\textbf{SlovakHateSpeechClassification.v2}} & \rotatebox{90}{\textbf{SlovakMovieReviewSentimentClassification.v2}} & \rotatebox{90}{\textbf{SlovakParlaSentClassification}} & \textbf{Avg} \\
\midrule 

\midrule
\multicolumn{9}{c}{\textit{Small models (<130M)}} \\
\rowcolor{gray!15} \texttt{e5-sk-small} \textsubscript{\textcolor{gray!60}{\tiny \textbf{(45M)}}} & 58.52 & 45.24 & 84.45 & 73.04 & 54.92 & 60.48 & 49.23 & 60.84 \\
\texttt{granite-embedding-107m-multilingual} \textsubscript{\textcolor{gray!60}{\tiny \textbf{(107M)}}} & 53.98 & 38.02 & 74.73 & 70.93 & 54.94 & 56.99 & 42.67 & 56.04 \\
\texttt{static-similarity-mrl-multilingual-v1} \textsubscript{\textcolor{gray!60}{\tiny \textbf{(108M)}}} & 51.72 & 48.49 & 56.86 & 41.86 & 48.82 & 55.33 & 39.02 & 48.87 \\
\texttt{multilingual-e5-small} \textsubscript{\textcolor{gray!60}{\tiny \textbf{(118M)}}} & 56.09 & 43.57 & 83.53 & 71.72 & 54.65 & 60.31 & 48.81 & 59.81 \\
\texttt{paraphrase-multilingual-MiniLM-L12-v2} \textsubscript{\textcolor{gray!60}{\tiny \textbf{(118M)}}} & 52.81 & 39.52 & 89.01 & 71.18 & 53.99 & 64.12 & 55.17 & 60.83 \\
\texttt{slovakbert-skquad-mnlr} \textsubscript{\textcolor{gray!60}{\tiny \textbf{(125M)}}} & 58.28 & 48.81 & 85.87 & 67.75 & 55.80 & 66.78 & 54.13 & 62.49 \\
\texttt{slovakbert-sts-stsb} \textsubscript{\textcolor{gray!60}{\tiny \textbf{(125M)}}} & 62.42 & 49.44 & 86.66 & 62.70 & 52.55 & 70.64 & 56.90 & 63.05 \\
\rowcolor{gray!15} \texttt{sturovec-base} \textsubscript{\textcolor{gray!60}{\tiny \textbf{(125M)}}} & 55.16 & 44.29 & 86.45 & 71.76 & 55.46 & 69.98 & 52.23 & 62.19 \\
\midrule
\multicolumn{9}{c}{\textit{Base models (>=130M, <350M)}} \\
\texttt{nomic-embed-text-v1.5} \textsubscript{\textcolor{gray!60}{\tiny \textbf{(137M)}}} & 57.58 & 39.52 & 74.31 & 42.70 & 51.62 & 55.78 & 38.27 & 51.40 \\
\texttt{granite-embedding-278m-multilingual} \textsubscript{\textcolor{gray!60}{\tiny \textbf{(278M)}}} & 56.95 & 42.14 & 76.76 & 70.64 & 55.02 & 59.09 & 44.08 & 57.81 \\
\texttt{multilingual-e5-base} \textsubscript{\textcolor{gray!60}{\tiny \textbf{(278M)}}} & 61.88 & 44.68 & 85.88 & 70.10 & 55.93 & 65.28 & 52.67 & 62.35 \\
\texttt{paraphrase-multilingual-mpnet-base-v2} \textsubscript{\textcolor{gray!60}{\tiny \textbf{(278M)}}} & 59.30 & 41.51 & 89.74 & 75.69 & 54.68 & 68.60 & \textbf{58.85} & 64.05 \\
\texttt{gte-multilingual-base} \textsubscript{\textcolor{gray!60}{\tiny \textbf{(305M)}}} & 57.42 & 43.49 & 85.14 & 75.74 & 53.19 & 65.72 & 51.21 & 61.70 \\
\texttt{embeddinggemma-300m} \textsubscript{\textcolor{gray!60}{\tiny \textbf{(308M)}}} & 53.59 & 46.43 & 88.83 & 71.32 & 53.48 & 72.15 & 48.77 & 62.08 \\
\texttt{nomic-embed-text-v2-moe} \textsubscript{\textcolor{gray!60}{\tiny \textbf{(330M)}}} & 60.86 & \textbf{49.84} & 84.96 & 76.08 & 54.37 & 70.99 & 50.79 & 63.98 \\
\midrule
\multicolumn{9}{c}{\textit{Large models (>=350M)}} \\
\rowcolor{gray!15} \texttt{e5-sk-large} \textsubscript{\textcolor{gray!60}{\tiny \textbf{(365M)}}} & \underline{63.12} & \underline{49.76} & 90.41 & 74.61 & 58.25 & 72.44 & 55.81 & 66.34 \\
\texttt{LaBSE} \textsubscript{\textcolor{gray!60}{\tiny \textbf{(471M)}}} & 59.22 & 46.90 & 84.17 & 57.11 & 57.34 & 59.44 & 46.90 & 58.73 \\
\texttt{multilingual-e5-large} \textsubscript{\textcolor{gray!60}{\tiny \textbf{(560M)}}} & 61.09 & 47.94 & 88.98 & 75.69 & 57.11 & 70.70 & 55.88 & 65.34 \\
\texttt{multilingual-e5-large-instruct} \textsubscript{\textcolor{gray!60}{\tiny \textbf{(560M)}}} & 62.81 & 46.51 & \underline{94.36} & \textbf{83.48} & \underline{59.86} & \underline{86.15} & \underline{58.77} & \underline{70.28} \\
\texttt{bge-m3} \textsubscript{\textcolor{gray!60}{\tiny \textbf{(568M)}}} & 58.59 & 45.48 & 93.24 & 72.55 & 57.71 & 81.97 & 58.10 & 66.81 \\
\texttt{snowflake-arctic-embed-l-v2.0} \textsubscript{\textcolor{gray!60}{\tiny \textbf{(568M)}}} & 58.91 & 44.76 & 89.74 & 75.00 & 57.96 & 67.58 & 49.87 & 63.40 \\
\texttt{jina-embeddings-v3} \textsubscript{\textcolor{gray!60}{\tiny \textbf{(572M)}}} & 57.27 & 45.00 & 93.48 & 76.23 & 55.16 & 83.54 & 57.58 & 66.89 \\
\texttt{Qwen3-Embedding-0.6B} \textsubscript{\textcolor{gray!60}{\tiny \textbf{(596M)}}} & 52.42 & 41.51 & 83.98 & \underline{79.31} & 55.55 & 76.25 & 50.83 & 62.84 \\
\texttt{jina-embeddings-v4} \textsubscript{\textcolor{gray!60}{\tiny \textbf{(3.8B)}}} & 60.08 & 44.52 & 87.19 & 77.89 & 55.33 & 66.83 & 49.12 & 62.99 \\
\texttt{Qwen3-Embedding-4B} \textsubscript{\textcolor{gray!60}{\tiny \textbf{(4B)}}} & 58.67 & 43.10 & 87.94 & 78.19 & 55.81 & 72.96 & 53.17 & 64.26 \\
\texttt{Qwen3-Embedding-8B} \textsubscript{\textcolor{gray!60}{\tiny \textbf{(8B)}}} & 62.03 & 44.21 & 87.67 & 78.92 & 56.94 & 76.50 & 55.33 & 65.94 \\
\midrule
\multicolumn{9}{c}{\textit{API access models}} \\
\texttt{embed-v4.0} & 57.66 & 45.48 & 81.02 & 72.75 & 54.79 & 62.42 & 46.73 & 60.12 \\
\texttt{text-embedding-3-small} & 57.11 & 47.78 & 83.46 & 74.41 & 56.35 & 67.34 & 48.46 & 62.13 \\
\texttt{text-embedding-3-large} & 61.09 & 47.70 & 90.62 & 79.17 & 56.87 & 79.64 & 53.31 & 66.91 \\
\texttt{gemini-embedding-001} & \textbf{66.64} & 48.97 & \textbf{94.99} & 75.78 & \textbf{62.83} & \textbf{91.08} & 56.77 & \textbf{71.01} \\
\texttt{amazon-titan-embed-text-v2} & 55.62 & 38.49 & 85.09 & 63.73 & 52.85 & 69.61 & 49.44 & 59.26 \\
\bottomrule
\end{tabular}
}
\caption{SkMTEB classification results (percent). Columns correspond to each classification dataset; the final \textbf{Avg} column is the unweighted mean across available classification tasks for a model. Models are grouped by size bucket (Small/Base/Large), followed by API-access models. The best result per task is \textbf{bolded} with the runner-up \underline{underlined}.}
\label{tab:mteb_results_clf}
\end{table*}
\end{document}

%% file: datasheet_table.tex
\begin{table*}[t]
\centering
\footnotesize
\setlength{\tabcolsep}{4pt}
\renewcommand{\arraystretch}{1.1}
\resizebox{\textwidth}{!}{%
\begin{tabular}{lllllr}
\toprule
\textbf{Dataset} & \textbf{Task} & \textbf{Domain} & \textbf{Origin} & \textbf{License} & \textbf{Splits (train/valid/test)} \\
\midrule
\multicolumn{6}{l}{\textit{Retrieval (5)}} \\
\midrule
\texttt{BelebeleRetrieval} & Rtrvl & Web, News & Native & CC-BY-SA-4.0 & --- / --- / 900 \\
\texttt{SKQuadRetrieval} & Rtrvl & Encyclopaedic & Native & CC-BY-NC-SA-4.0 & --- / --- / 1{,}134 \\
\texttt{SlovakSumRetrieval} & Rtrvl & News, Social & Native & OpenRAIL & --- / --- / 600 \\
\texttt{SMESumRetrieval} & Rtrvl & News, Social & Native & N/A & --- / --- / 600 \\
\texttt{WebFAQRetrieval} & Rtrvl & Web & Native & CC-BY-4.0 & --- / --- / 3{,}153 \\
\midrule
\multicolumn{6}{l}{\textit{Reranking (3)}} \\
\midrule
\texttt{SkQuadReranking} & Rrnk & Encyclopaedic & Native & CC-BY-SA-4.0 & --- / --- / 1{,}133 \\
$\star$\,\texttt{SlovakPharmacyDrMaxReranking} & Rrnk & Medical, Web & Author-created & CC-BY-NC-ND-4.0 & --- / --- / 4{,}676 \\
$\star$\,\texttt{SlovakPharmacyMojaLekarenReranking} & Rrnk & Medical, Web & Author-created & CC-BY-NC-ND-4.0 & --- / --- / 738 \\
\midrule
\multicolumn{6}{l}{\textit{Classification (7)}} \\
\midrule
\texttt{MultiEupSlovakGenderClassification} & Clf & Government & Native & CC-BY-4.0 & 508 / --- / 128 \\
\texttt{MultiEupSlovakPartyClassification} & Clf & Government & Native & CC-BY-4.0 & 502 / --- / 126 \\
\texttt{MultilingualSentimentClassification} & Clf & Reviews & Native & N/A & 3{,}560 / 522 / 1{,}042 \\
\texttt{SIB200Classification} & Clf & News & Native & CC-BY-SA-4.0 & 701 / 99 / 204 \\
\texttt{SlovakHateSpeechClassification.v2} & Clf & Social & Native & CC-BY-SA-4.0 & 11{,}301 / --- / 1{,}237 \\
\texttt{SlovakMovieReviewSentimentClassification.v2} & Clf & Reviews & Native & CC-BY-NC-SA-4.0 & 20{,}181 / 2{,}083 / 2{,}008 \\
\texttt{SlovakParlaSentClassification} & Clf & Government & Native & CC-BY-SA-4.0 & 2{,}080 / --- / 520 \\
\midrule
\multicolumn{6}{l}{\textit{Clustering (5)}} \\
\midrule
$\star$\,\texttt{PravdaSKTagClustering} & Clust & News & Author-created & N/A & --- / --- / 15{,}000 \\
$\star$\,\texttt{PravdaSKURLClustering} & Clust & News & Author-created & N/A & --- / --- / 15{,}000 \\
\texttt{SIB200ClusteringS2S} & Clust & News & Native & CC-BY-SA-4.0 & --- / --- / 204 \\
\texttt{SMESumCategoryClustering} & Clust & News & Native & N/A & --- / --- / 7{,}233 \\
\texttt{SlovakSumURLClustering} & Clust & News & Native & N/A & --- / --- / 10{,}871 \\
\midrule
\multicolumn{6}{l}{\textit{Bitext Mining (6)}} \\
\midrule
\texttt{FloresBitextMining} & Btxt & Non-fiction, Encyclopaedic & Native & CC-BY-SA-4.0 & --- / --- / 1{,}012 \\
\texttt{NTREXBitextMining} & Btxt & News & Native & CC-BY-SA-4.0 & --- / --- / 1{,}997 \\
\texttt{OpusSlovakEnglishBitextMining} & Btxt & Web, Subtitles & Native & N/A & 1{,}000{,}000 / 2{,}000 / 2{,}000 \\
\texttt{Tatoeba} & Btxt & Written & Native & CC-BY-2.0 & --- / --- / 1{,}000 \\
\texttt{WebFAQBitextMiningQAs} & Btxt & Web & Native & CC-BY-4.0 & --- / --- / 2{,}551+1{,}823 \\
\texttt{WebFAQBitextMiningQuestions} & Btxt & Web & Native & CC-BY-4.0 & --- / --- / 2{,}551+1{,}823 \\
\midrule
\multicolumn{6}{l}{\textit{Pair Classification (3)}} \\
\midrule
$\star$\,\texttt{DemagogSKNLI} & PrClf & Government, News & Author-created & N/A & --- / --- / 3{,}085 \\
$\star$\,\texttt{SlovakNLI} & PrClf & News, Web & Author-created & N/A & --- / --- / 382 \\
\texttt{SlovakRTE} & PrClf & News, Web & Translated & CC-BY-SA-4.0 & 2{,}490 / 277 / 1{,}660 \\
\midrule
\multicolumn{6}{l}{\textit{STS (2)}} \\
\midrule
\texttt{SlovakSTS} & STS & News, Blog & Translated & CC-BY-SA-4.0 & 5{,}604 / 1{,}481 / 1{,}352 \\
$\star$\,\texttt{SlovakSumSTS} & STS & News & Author-created & CC-BY-NC-4.0 & 4{,}926 / --- / 298 \\
\bottomrule
\end{tabular}%
}
\caption{Datasheet for the 31 SkMTEB datasets. \textbf{Origin}: Native = natively Slovak; Translated = translated from another language; Author-created ($\star$) = introduced in this work. \textbf{Splits} lists the sizes of the train / valid / test splits as provided by the task definition; ``---'' means the split is not provided. SkMTEB evaluation uses the test split in every case. \textbf{Task} codes: Rtrvl = Retrieval, Rrnk = Reranking, Clf = Classification, Clust = Clustering, Btxt = Bitext Mining, PrClf = Pair Classification, STS = Semantic Textual Similarity. For bitext mining, test sizes are per language pair; Flores and NTREX are evaluated on both eng--slk and ces--slk pairs, WebFAQ variants list both pair sizes (eng--slk / ces--slk), and Tatoeba and Opus use slk--eng only.}
\label{tab:datasheet}
\end{table*}